\documentclass[10pt,twocolumn,letterpaper]{article}

\usepackage{iccv}
\usepackage{times}
\usepackage{epsfig}
\usepackage{graphicx}
\usepackage{amsmath}
\usepackage{amssymb}

\usepackage[accsupp]{axessibility}  
\usepackage{booktabs}
\usepackage{blindtext}
\usepackage{soul}
\usepackage{xcolor}
\usepackage{multirow}

\def\secref#1{Section~\ref{#1}}
\def\figref#1{Figure~\ref{#1}}
\def\tabref#1{Table~\ref{#1}}
\def\eqref#1{Equation~(\ref{#1})}

\newcommand{\name}[0]{3DMOTFormer}

\usepackage[breaklinks=true,bookmarks=false]{hyperref}

\iccvfinalcopy 

\def\iccvPaperID{7435} 

\ificcvfinal\fi

\begin{document}

\title{3DMOTFormer: Graph Transformer for Online 3D Multi-Object Tracking}

\author{Shuxiao Ding$^{1,2}$, \quad Eike Rehder$^{3}$, \quad Lukas Schneider$^{1}$, \quad Marius Cordts$^{1}$, \quad Juergen Gall$^{2,4}$\\
$^1$Mercedes-Benz AG, Sindelfingen, Germany, \\
$^2$University of Bonn, Bonn, Germany, \\
$^3$Robert Bosch GmbH, Stuttgart, Germany \\
$^4$Lamarr Institute for Machine Learning and Artificial Intelligence, Germany \\
{\tt\small \{shuxiao.ding,lukas.schneider,marius.cordts\}@mercedes-benz.com}, \\
{\tt\small e.rehder@gmx.de}, \quad {\tt\small gall@iai.uni-bonn.de}
}

\maketitle
\ificcvfinal\thispagestyle{empty}\fi

\begin{abstract}
  Tracking 3D objects accurately and consistently is crucial for autonomous vehicles, enabling more reliable downstream tasks such as trajectory prediction and motion planning.
  Based on the substantial progress in object detection in recent years, the tracking-by-detection paradigm has become a popular choice due to its simplicity and efficiency.
  State-of-the-art 3D multi-object tracking (MOT) approaches typically rely on non-learned model-based algorithms such as Kalman Filter but require many manually tuned parameters.
  On the other hand, learning-based approaches face the problem of adapting the training to the online setting, leading to inevitable distribution mismatch between training and inference as well as suboptimal performance.
  In this work, we propose \name{}, a learned geometry-based 3D MOT framework building upon the transformer architecture.
  We use an Edge-Augmented Graph Transformer to reason on the track-detection bipartite graph frame-by-frame and conduct data association via edge classification.
  To reduce the distribution mismatch between training and inference, we propose a novel online training strategy with an autoregressive and recurrent forward pass as well as sequential batch optimization.
  Using CenterPoint detections, our approach achieves 71.2\% and 68.2\% AMOTA on the nuScenes validation and test split, respectively.
  In addition, a trained \name{} model generalizes well across different object detectors.
  Code is available at: \url{https://github.com/dsx0511/3DMOTFormer}.

\end{abstract}

\section{Introduction}
\label{sec:intro}

\begin{figure}[h]
  \begin{center}
    \resizebox{\linewidth}{!}{
      \includegraphics[width=\linewidth]{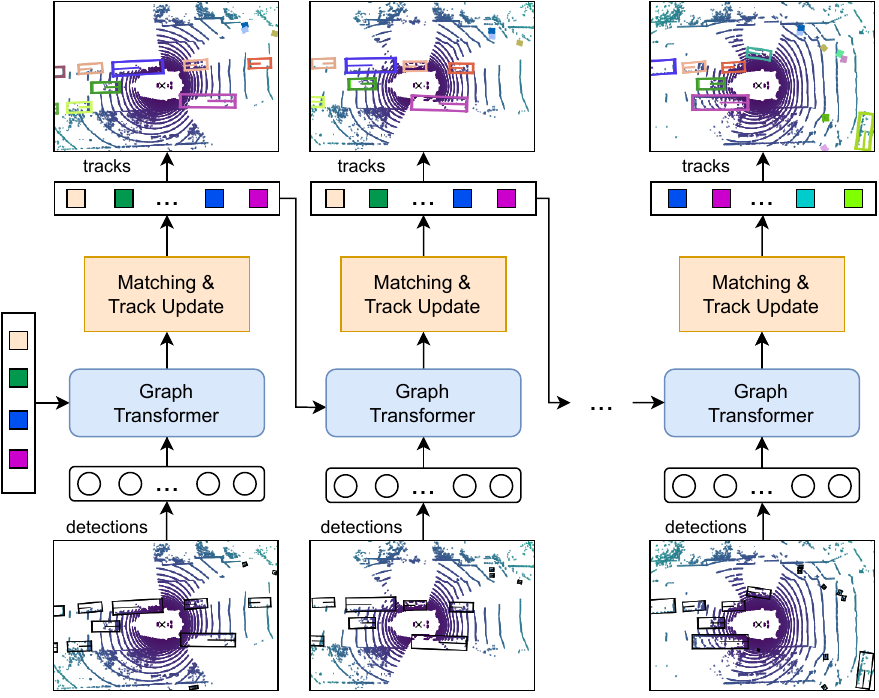}
    }
  \end{center}
  \caption{We propose \name{} that reasons on the track-detection bipartite graph and estimates the data association using a Graph Transformer. Tracks for the next frame are generated by the matching and track update module autoregressively.}
  \label{fig:title}
\end{figure}

3D multi-object tracking (MOT) is a fundamental task in many applications such as autonomous driving and mobile robots, aiming at localization, classification and persistent identification of surrounding objects over time.
Especially accurate and consistent online tracking is of great importance for downstream tasks such as trajectory prediction, motion planning and robot navigation.

Due to recent advances in object detection performance, the tracking-by-detection paradigm has become a popular choice to accomplish MOT~\cite{Bewley2016SimpleOA,Wojke2017SimpleOA,Zhang2021ByteTrackMT,Weng20193DMT,Yin2020Centerbased3O,Chiu2020Probabilistic3M}.
Most tracking-by-detection approaches utilize detections from every frame generated with an off-the-shelf object detector and focus on associating the detection results across frames.

State-of-the-art tracking-by-detection methods typically use non-learned algorithms, \eg Kalman Filters~\cite{Weng20193DMT,Chiu2020Probabilistic3M,Benbarka2021ScoreRF,Kim2021EagerMOT3M}, with a pre-defined motion assumption, \eg constant turn rate and velocity (CTRV) model, followed by a geometric association metric, \eg center distance or 3D IoU, which requires a lot of handcrafting and heuristics.
Learning-based approaches on the other hand aim at reducing heuristics but face the challenge of lifting the training to the online inference setting.
Some approaches~\cite{Chu2021TransMOTSG,Weng2020GNN3DMOTGN,Kim2022PolarMOTHF} adopt a teacher-forcing~\cite{Williams1989ALA} training using ground truth trajectories with annotated instance IDs and/or annotated bounding boxes.
However, during online inference, the network has to associate noisy detections to the tracked trajectories containing false associations caused by the network itself.
This results in a strong distribution mismatch or over-fitting despite applying plenty of data augmentations.
Another line of works~\cite{Braso2019LearningAN,Zaech2021LearnableOG,Kim2022PolarMOTHF} regards detections as nodes in a spatiotemporal graph and applies Neural Message Passing (NMP)~\cite{Gilmer2017neural}.
However, for online MOT, these methods require a graph with a fixed time window to evolve frame-by-frame, while the training is done on the static graph with the same time window but without dynamic evolving.
OGR3MOT~\cite{Zaech2021LearnableOG} worked on adapting this graph representation to the online setting but still uses a semi-online training and uses an additional heuristic track update for inference.

In this work, we present a novel transformer-based 3D MOT framework, which we call \name{}, that is learnable and relies only on geometric cues, as shown in~\figref{fig:title}.
Our model iteratively reasons on the relationship between existing tracks as well as detections in a new frame and conducts association using edge classification.
A greedy matching and a simple track update module generate tracks as input for the next frame, yielding an autoregressive loop.
This results in a bipartite graph representation between tracks and detections.
In contrast to existing approaches that process a spatiotemporal graph with a fixed time window~\cite{Zaech2021LearnableOG,Kim2022PolarMOTHF}, we directly feed the processed track features into the new frame as initial track features to access temporal information, similar to the hidden states in RNNs.
To tackle the different operation modes between training and test time, we propose a novel fully online training strategy which consists of an \textit{autoregressive forward pass} and a \textit{sequential batch backward pass}.
Concretely, identical to the inference phase, our model evolves frame-by-frame autoregressively on sampled sequence clips during training, instead of modelling the sequence in a graph as a whole.
We accumulate the loss at each frame and optimize the network after the whole training sequence was processed.
The forward pass fully simulates the operation mode and the data distribution during the online inference phase, while the optimization method learns to recover from errors and considers the whole sequence.
Considering the remarkable achievements of autoregressive models based on transformers in natural language processing~\cite{Vaswani2017attention,Radford2018ImprovingLU,Brown2020LanguageMA},
we use Edge-Augmented Graph Transformers~\cite{Hussain2021EdgeaugmentedGT}, a variant of transformers that generalizes to sparse graphs and takes edge features into account for attention calculation.
Also, structural information in the bipartite graphs between tracks and detections can be effectively captured using cross-attention, which justifies transformer-based models as a suitable choice for our MOT framework.

We evaluate our method on the nuScenes~\cite{Caesar2019nuScenesAM} tracking benchmark using CenterPoint detections~\cite{Yin2020Centerbased3O} as input.
Our method achieves 71.2\% and 68.2\% AMOTA on the validation and test split, respectively, yielding state-of-the-art performance among all geometry-based approaches.
We show the generalization of \name{} where a frozen \name{} model still achieves competitive performance when inferring on detections from another detector.

Our main contributions can be summarized as follows:
\begin{itemize}
  \item We propose \name{}, a novel online 3D MOT framework based on Edge Augmented Graph Transformers~\cite{Hussain2021EdgeaugmentedGT} for learning data association, which reduces the need for handcrafted components compared to previous state-of-the-art.
  \item \name{} is tailored towards an online training strategy for MOT, which fully mimics the setup and hence the data distribution during online inference.
  \item Our method achieves state-of-the-art performance, in particular 71.2\% and 68.2\% AMOTA on the nuScenes~\cite{Caesar2019nuScenesAM} validation and test split, respectively, using CenterPoint detections~\cite{Yin2020Centerbased3O} as input.
  \item \name{} achieves competitive performance when inferring on detections from another detector. This allows flexible deployment of the same \name{} model independent of the object detector.
\end{itemize}

\section{Related Work}
\label{sec:related}

In this section, we discuss previous works based on the tracking-by-detection paradigm with a focus on 3D MOT.
These works can be divided into model-based and learning-based approaches.
Some approaches adopt neural networks as a complementary module in their pipeline but still heavily rely on the pre-defined motion model.
Thus, we still categorized them into model-based approaches.

\subsection{Model-based multi-object tracking}
Using Kalman Filters (KF) with a 2D motion model achieved great success in 2D MOT, \eg SORT~\cite{Bewley2016SimpleOA} with its succeeding variants~\cite{Wojke2017SimpleOA,Du2022StrongSORTMD,Aharon2022BoTSORTRA} and ByteTrack~\cite{Zhang2021ByteTrackMT}.
Inspired by this success, AB3DMOT~\cite{Weng20193DMT} uses a 3D motion model for KF and an association using 3D IoU.
Some works improve AB3DMOT using other association metrics, \eg Mahalanobis distance in Chiu \etal~\cite{Chiu2020Probabilistic3M} and 3D Generalized IoU (GIoU)~\cite{Rezatofighi2019GeneralizedIO} in SimpleTrack~\cite{Pang2021SimpleTrackUA}.
Others investigate the track life management module, \eg the confidence-based track spawn and termination in CBMOT~\cite{Benbarka2021ScoreRF} and the permanent preservation without termination in ImmortalTracker~\cite{Wang2021ImmortalTT}.
CenterTrack~\cite{Zhou2020TrackingOA} and CenterPoint~\cite{Yin2020Centerbased3O} replace the filter algorithm with a simpler constant velocity model where they use a temporal object detection model with accurate velocity regression.
Many recent 3D object detection works~\cite{Bai2022TransFusionRL,jiao2022msmdfusion,Li2022UnifyingVR,Liu2022BEVFusionMM} achieve competitive tracking performance using the CenterPoint-based tracker.
GNN-PMB~\cite{Liu2022GNNPMBAS} proposes a Poisson multi-Bernoulli filter using global nearest neighbor for data association and achieves state-of-the-art performance.
Although model-based approaches for MOT so far performed better than learned ones, our learning-based method is able to outperform model-based methods while reducing the need for manual parameter tuning and designing heuristics.
Another line of works incorporates additional data or multi-modal sensor fusion, including both geometric and appearance cues.
EagerMOT~\cite{Kim2021EagerMOT3M} fuses 3D and 2D detections on a greedy basis in a two-stage association.
Other works use a CNN-based image feature extractor and conduct appearance-based association complementary to geometric metrics, \eg Chiu \etal~\cite{Chiu2020Probabilistic3MM} and CAMO-MOT~\cite{Wang2022CAMOMOTCA}.

\subsection{Learning-based multi-object tracking}
Learning-based methods usually use a Graph Neural Network (GNN) to address the association task.
The first group of works~\cite{Weng2020GNN3DMOTGN,Chu2021TransMOTSG} treats object association as a bipartite graph between tracked trajectories and detections.
They typically use a temporal encoding of the tracks, \eg LSTM in GNN3DMOT~\cite{Weng2020GNN3DMOTGN} or spatiotemporal Transformer in TransMOT~\cite{Chu2021TransMOTSG}.
As a result, an offline teacher-forcing~\cite{Williams1989ALA} training with ground truth object associations is needed, which leads to overfitting despite data augmentations.
Another group of works~\cite{Braso2019LearningAN,Rangesh2021TrackMPNNAM,Kim2022PolarMOTHF,Zaech2021LearnableOG} uses a spatiotemporal graph with a temporal window size where the association is done for every consecutive frame.
In 2D MOT, MPNTrack~\cite{Braso2019LearningAN} uses a Message Passing Network (MPN)~\cite{Gilmer2017neural} to address the offline tracking as a min-cost network flow problem~\cite{Zhang2008GlobalDA}.
TrackMPNN~\cite{Rangesh2021TrackMPNNAM} moves MPNTrack~\cite{Braso2019LearningAN} towards the online setting by updating the graph dynamically as a rolling window and accumulating losses over the sequence during training.
In 3D MOT, OGR3MOT~\cite{Zaech2021LearnableOG} lifts MPNTrack~\cite{Braso2019LearningAN} to the online setting by extending predictive track nodes based on KF and a semi-online training method.
This training method consists of two stages: the first stage uses ground truth to generate track node data, whereas in the second stage, the track node data is inferred using a trained first-stage model.
However, these works need a complex heuristic algorithm to decode multi-frame network outputs into associated trajectories~\cite{Rangesh2021TrackMPNNAM,Zaech2021LearnableOG} which needs to resolve conflicts between different track hypotheses.
In contrast, our method uses a dynamic bipartite graph, which requires a much simpler track update operation to decode the network output into hard association.
This enables a fully autoregressive forward pass during training to further approximate the online inference.
We use the accumulated loss over multiple frames to train the network, similar to~\cite{Rangesh2021TrackMPNNAM}, in order to extend the temporal receptive field of the bipartite graph and to optimize the online inference process.

\begin{figure*}[t!]
  \begin{center}
    \includegraphics[width=\textwidth]{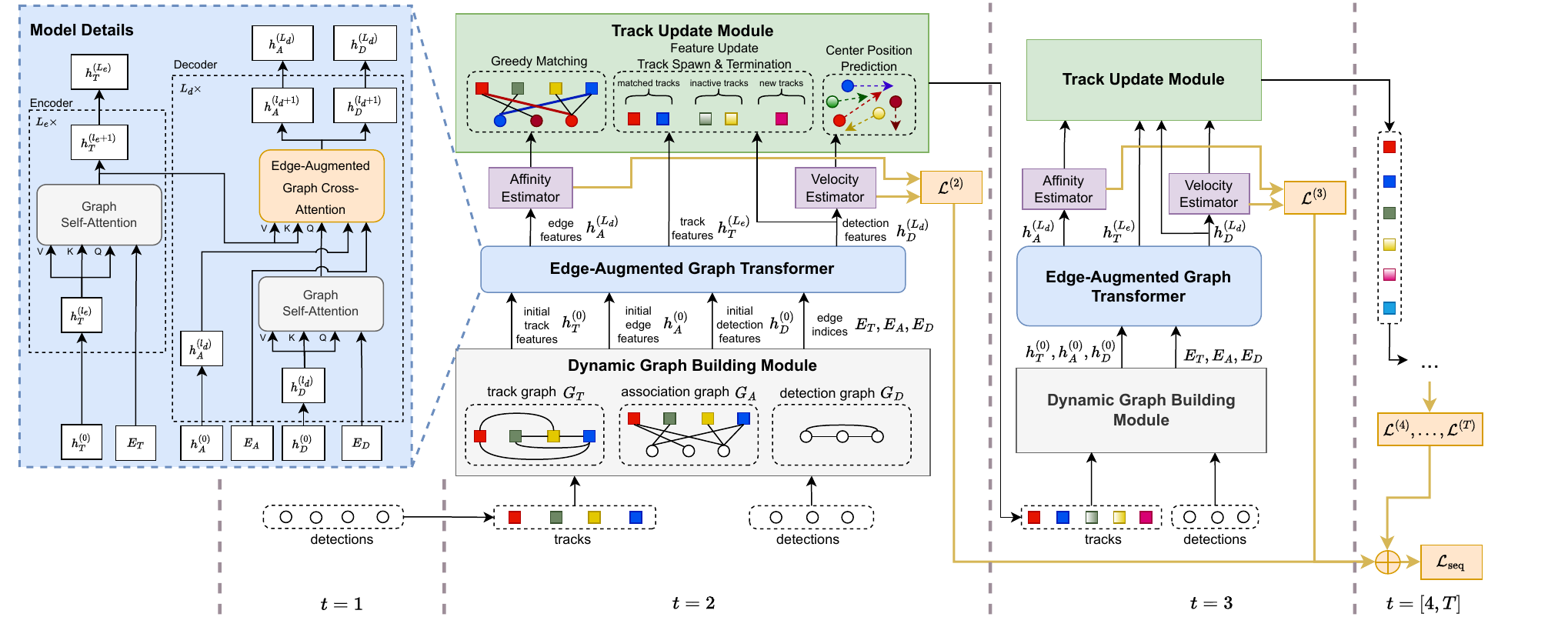}
  \end{center}
  \vspace{-2mm}
  \caption{Overview of our \name{} framework. Our model processes graph-structured data consisting of tracks and detections. We use an encoder with graph self-attention to encode existing tracks, and a decoder with both graph self-attention and edge-augmented graph cross-attention~\cite{Hussain2021EdgeaugmentedGT} that processes detection and edge features. The decoder outputs are used to estimate affinity and velocity, which are further used to update tracks. The network runs autoregressively during training and we optimize the network using the losses of all frames.%
  }
  \label{fig:overview}
\end{figure*}

\section{Our Approach}
\label{sec:main}

An overview of our proposed \name{} is shown in~\figref{fig:overview}.
Given existing tracks and new detections at time stamp $t$, we first build track $G_T$, detection $G_D$ and association graphs $G_A$.
The initial track $h_T^{(0)}$, detection $h_D^{(0)}$, and edge features $h_A^{(0)}$ as well as the graph structure are processed by a transformer-based model consisting of graph self-attention and edge-augmented graph cross-attention~\cite{Hussain2021EdgeaugmentedGT}.
Based on updated edge features $h_A^{(L_d)}$, we compute the affinity between tracks and detections which is further processed by the track update module to autoregressively generate inputs for the next frame.
We also estimate velocities to predict track locations.
Our training stage fully mimics this online inference schema:
we run our model in an autoregressive manner, calculate the affinity and velocity loss for each frame, and finally propagate gradients after the entire training sequence has been processed.

\subsection{Graph Representation}
\label{sec:graph_repr}
We model the multi-object tracking using a graph representation, where a detection or a tracklet is regarded as a node.
We employ a sparse graph representation to remove redundant connections and provide more structured data for a better interaction modelling and feature extraction.
We build three graphs:
(1) a detection graph $G_D$ which enables a message passing between detections to update detection features with a scene embedding,
(2) a track graph $G_T$ which models the interactions between existing unassociated tracks from previous frames,
and (3) an association graph $G_A$ which reasons on the relationship between tracks and detections that will be potentially associated.

\paragraph{Detection graph}
The detection graph $G_D = (V_D, E_D)$ with detection nodes $V_D$ and detection edges $E_D$ is built to model the interaction between newly detected objects, where each object represents a node $v_{D,i} \in V_D$.
Similar to OGR3MOT~\cite{Zaech2021LearnableOG}, the initial feature embedding of a detection node $x_{D,i}$ is a concatenation of box center position, size, yaw, velocity, one-hot encoded class and detection score. All these values are readily available in typical 3D detectors, \eg CenterPoint~\cite{Yin2020Centerbased3O}. Next,
$x_{D,i}$ is embedded using a Multi-Layer Perceptron (MLP) into $h^{(0)}_{D,i}$ as the input of the model.
The detection graph is truncated using a fixed class-agnostic distance threshold.

\paragraph{Track graph}
The track graph $G_T = (V_T, E_T)$ with track nodes $V_T$ and track edges $E_T$ is similar to the detection graph $G_D$.
As our model runs recurrently, for each track in the memory, we use the processed track feature from the previous frame as the initial feature $h_{T,i}^{(0)}$ in a new frame.
This initial feature is similar to the hidden state in an RNN, which enables the model to access the information from the history.
The track edges $E_T$ are established with the same truncation mechanism as in the detection graph.

\vspace{-2mm}
\paragraph{Association graph}
The association graph $G_A = (V_D, V_T, E_A)$ is a bipartite graph that connects detections $V_D$ and tracks $V_T$ using association edges $E_A$.
We first predict the position of all tracks in the new frame assuming constant velocity, where the velocity is estimated by our network.
In contrast to the other two graphs, an association edge is only established between two nodes with the same category and it is truncated using class-specific distance thresholds.
The thresholds are calculated based on the dataset statistic of the maximal velocity of a certain class, following~\cite{Yin2020Centerbased3O}.
In addition, every association edge $e_{A,ij} \in E_A$ contains a corresponding initial edge feature $h_{A,ij}^{(0)}$.
We follow OGR3MOT~\cite{Zaech2021LearnableOG} to use original position difference, size difference, yaw difference, frame difference and the center distance after prediction.
Using edge features, we enable a more comprehensive interaction modelling between tracks as well as detections.

\subsection{Graph Transformer for MOT}
Transformers~\cite{Vaswani2017attention} are widely used as an autoregressive model in natural language processing.
Also, they have shown promising performance on soft association, \eg for feature matching~\cite{Sarlin2019SuperGlueLF,Sun2021LoFTRDL} and for multi-modal sensor fusion~\cite{Bai2022TransFusionRL}.
Our work also runs autoregressively and the soft association with attention-weighted feature aggregation can help to implicitly acquire multiple hypotheses from past frames, which makes transformers a suitable choice for MOT.
The following sections present the adaptation of transformers to our graph representation.

\vspace{-2mm}
\paragraph{Graph Transformer encoder}
As shown in the top left side of~\figref{fig:overview}, the Graph Transformer encoder generates updated feature encodings of existing tracks by modelling interaction between them using self-attention.
For each layer $l$, the multi-head attention uses three different linear layers to project the track node features $h^{(l)}_{T,i}$ into value $v^{(l)}_{ic}$, key $k^{(l)}_{ic}$ and query $q^{(l)}_{ic}$, where $c$ is an index for attention heads with $c \in [1, C]$.
For each head, the attention from $j$-th to $i$-th node is calculated using inner product between key and query $\langle q,k \rangle = \frac{q^{T} k}{\sqrt{d}}$, divided by $\sqrt d$ where $d$ denotes the model feature dimension.
A normalized attention $\alpha_{ijc}^{(l)}$ is calculated by
\vspace{-2mm}
\begin{equation}
  \alpha_{ijc}^{(l)} = \frac{\exp\big(\langle q^{(l)}_{ic}, k^{(l)}_{jc} \rangle\big)}{\sum_{m \in \mathcal{N}(i)}  \exp\big(\langle q^{(l)}_{ic}, k^{(l)}_{mc} \rangle\big)},
  \label{equ:attn}
\end{equation}
where $\mathcal{N}(i)$ denotes the neighbors of node $i$ and it is defined by the graph connectivity (see~\secref{sec:graph_repr}).
~\eqref{equ:attn} corresponds to a sparse version of softmax that normalizes over neighbors of the node $i$ rather than all nodes.
The track feature is then updated by an attention-weighted aggregation over the value vectors of all neighboring nodes, followed by a concatenation of all attention heads:
\begin{equation}
  \hat{h}^{(l+1)}_{T,i} = W_{\text{O}}^{(l)} \Big(\mathbin\Big\Vert_{c=1}^C ( \sum_{j \in \mathcal{N}(i)} \alpha_{ijc}^{(l)} v^{(l)}_{jc}) \Big), 
  \label{equ:multi_heads}
\end{equation}
where $\mathbin\Vert$ is the concatenation operation and $W_{\text{O}}^{(l)}$ denotes learnable weights of a linear layer.
This self-attention uses a graph structure from edge indices $E$ to derive the neighborhood $\mathcal{N}(i)$ for each node $i$, making it different to a standard transformer that considers all other nodes.
$\hat{h}^{(l+1)}_{T,i}$ is processed by a Feed Forward Network (FFN) to generate the output of an encoder layer: $h^{(l+1)}_{T,i} {=} \text{FFN}(\hat{h}^{(l+1)}_{T,i})$.
The final track features $h^{(L_e)}_{T,i}$ are calculated by stacking $L_e$ layers.

\paragraph{Edge-Augmented Graph Transformer decoder}
Our decoder layer consists of a graph self-attention and a graph edge-augmented cross-attention~\cite{Hussain2021EdgeaugmentedGT}.
The self-attention is implemented in the same way as in the Graph Transformer encoder for new detections, which takes detection features $h^{(l)}_{D,i}$ and detection edges $E_D$ as input and produces intermediate detection features $\tilde{h}^{(l)}_{D,i}$.
At every decoder layer $l$, we project detection feature $\tilde{h}^{(l)}_{D,i}$ for each node $i$ into the query, while key and value are converted from the encoder output $h^{(L_e)}_{T,i}$ for all decoder layers.
Following~\cite{Hussain2021EdgeaugmentedGT}, besides the inner product of key and query, edge features $h^{(l)}_{A,ij}$ also contribute to the attention calculation, \ie
\begin{equation}
  \alpha_{ijc}^{(l)} = \frac{\exp\big(\langle q^{(l)}_{ic}, k^{(l)}_{jc} \rangle + W^{(l)}_{A,ijc} h^{(l)}_{A,ij}\big)}{\sum_{m \in \mathcal{N}(i)}  \exp\big(\langle q^{(l)}_{ic}, k^{(l)}_{mc} \rangle + W^{(l)}_{A,imc} h^{(l)}_{A,im}\big)},
\end{equation}
where $W^{(l)}_{A,ijc} \in \mathbb{R}^{d \times 1}$ projects the edge feature into a scalar which represents a part of the attention.
Same for the encoder, the output detection feature $h^{(l+1)}_{D,i}$ of every node $i$ is calculated using~\eqref{equ:multi_heads} and an FFN.
As for the edge features, we concatenate the cross-attentions before the softmax normalization of all heads and then project it back to the latent dimension $d$ using learnable weights $W_{O}^{(l)} \in \mathbb{R}^{c \times d}$, \ie
\begin{equation}
  \hat{h}^{(l+1)}_{A,ij} = W_{O}^{(l)} \Big(\mathbin\Big\Vert_c \big( \langle q^{(l)}_{ic}, k^{(l)}_{jc} \rangle + W^{(l)}_{A,ijc} h^{(l)}_{A,ij} \big)\Big).
\end{equation}
The final output edge feature $h^{(l+1)}_{A,ij}$ of layer $l$ is generated using another FFN: $h^{(l+1)}_{A,ij} = \text{FFN}(\hat{h}^{(l+1)}_{A,ij})$.
We also stack multiple decoder layers, resulting in the final detection features $h^{(L_d)}_{D,i}$ and final association edge features $h^{(L_d)}_{A,ij}$.

\subsection{Learning Targets}
\label{sec:learning_target}
\paragraph{Affinity estimation}
Based on the final edge feature $h^{(L_d)}_{A,ij}$, we use an MLP to estimate an affinity score $a_{ij} = \text{MLP}(h^{(L_d)}_{A,ij})$, which represents the probability that detection $i$ and track $j$ represent the same object.
As the detections are estimated from a 3D detector, we run a Hungarian Matching~\cite{Kuhn1955TheHM} between annotated and detection boxes using 3D Intersection-over-Union (IoU) as matching cost in order to assign a ground truth ID for detection boxes.
Unmatched detection boxes are marked as false positive.
The classification target of the edge of $e_{A,ij}$ is positive, only if its connected detection $i$ and track $j$ share the same ID.

\paragraph{Velocity estimation}
Although many state-of-the-art 3D detectors are able to estimate velocities, the estimation can be more accurate when the objects are tracked for a longer time.
In our framework, detections extract features from the hidden states of all tracks using cross-attention, thus capturing abundant historical information.
For each detection $i$, we use an MLP to regress its velocity of the box center $v_i = \text{MLP}(h^{(L)}_{D,i})$.
The ground truth velocity is generated for true positive detections using their ground truth annotations.
False positive detections are ignored during training.
After the track update module, we predict the positions of tracked objects in the next frames using their corresponding velocity $v_i$.
The predicted position is used in the association graph building as described in~\secref{sec:graph_repr}.

\paragraph{Loss function}
We use the Focal Loss~\cite{Lin2017FocalLF} with $\alpha=0.5$ and $\gamma=1.0$ as association loss $\mathcal{L}_\text{a}$ and smooth-$\ell_{1}$ loss as the velocity loss $\mathcal{L}_\text{v}$.
The overall loss is $\mathcal{L} = \mathcal{L}_\text{a} + \lambda_\text{v} \mathcal{L}_\text{v}$ where we set $\lambda_\text{v} = 1.0$ as default in our experiments.
We evaluate the impact of $\lambda_\text{v}$ is in the supplementary material.

\subsection{Fully Online Training}
\label{sec:online_training}
Unlike existing learned trackers~\cite{Weng2020GNN3DMOTGN,Zaech2021LearnableOG,Kim2022PolarMOTHF} that use ground truth trajectories including annotated IDs as input, we use those IDs only for calculating the loss.
However, an accurate association heavily relies on an observation of the motion in the past few frames.
We tackle this problem by generating trajectories autoregressively during training and optimize the network using sequential outputs as a whole.

The affinity score reflects a soft association but not a hard decision as required for the track update.
Hence, we use greedy bipartite matching, where we greedily match the detections starting with the highest detection score to the track with the highest affinity score, while a track cannot be matched twice.
For a matched detection-track pair, the track feature will be replaced by the matched detection feature.
We refer to the supplementary material for a detailed illustration of the track feature update.
Based on the matching results, a simple heuristic track life management determines track spawning and termination.
To achieve high recall, we initialize all unmatched detections as new tracks.
We use a count-based track management where all tracks that are unmatched for $T_d$ frames are permanently deleted.
The unmatched tracks with an age smaller than $T_d$ are temporally inactive but still used in the next frame.

With the track update module in the loop, we accomplish a frame-by-frame autoregressive forward pass during training, identical to online inference.
While we generate trajectories using the network itself, it can introduce errors which could significantly affect the training of subsequent frames.
To solve this problem, our training strategy aims at optimizing the network using the whole sequence instead of each frame.
Concretely, we store the losses for each time stamp $\mathcal{L}^{(t)}$ while processing the training sequence frame-by-frame.
When the whole training sequence of length $T$ has been processed by the network, we accumulate the losses of each time stamp to get the sequence loss: $\mathcal{L}_\text{seq} = \sum_{t=2}^T \mathcal{L}^{(t)}$.
We then execute the back-propagation through time (BPTT)~\cite{Rumelhart1986LearningRB} to optimize the network using losses from all time stamps.
Using this sequential batch optimization method, the network is trained to capture and correct the errors in the previous frames, which subsequently brings a better online performance.


\begin{table*}[t!]
  \centering
  \small
  \setlength{\tabcolsep}{3pt}

  \begin{center}
    \begin{tabular}{lccccccccccccc}
      \toprule
      Method                                              & Additional Cues &  & AMOTA$\uparrow$   & AMOTP$\downarrow$ &  & MOTA$\uparrow$    & MT$\uparrow$  & ML$\downarrow$   & TP$\uparrow$   & FP$\downarrow$ & FN$\downarrow$    & IDS$\downarrow$ & FRAG$\downarrow$ \\
      \midrule
      OGR3MOT~\cite{Zaech2021LearnableOG}                 & --              &  & 0.656             & 0.620             &  & 0.554             & 5278          & 2094             & 95264          & 17877          & 24013             & 288             & 371              \\
      PolarMOT-offline$^\dagger$~\cite{Kim2022PolarMOTHF} & --              &  & 0.664             & 0.566             &  & 0.561             & \textbf{5701} & 1686             & \textbf{97909} & 17856          & \textbf{21414}    & \textbf{242}    & \textbf{332}     \\
      CenterPoint~\cite{Yin2020Centerbased3O}             & --              &  & 0.638             & 0.555             &  & 0.537             & 5584          & 1681             & 95877          & 18612          & 22928             & 760             & 529              \\
      CBMOT~\cite{Benbarka2021ScoreRF}                    & --              &  & 0.649             & 0.592             &  & 0.545             & 5319          & 1966             & 94916          & \textbf{16469} & 24092             & 557             & 450              \\
      SimpleTrack$^\ddagger$~\cite{Pang2021SimpleTrackUA} & --              &  & 0.668             & 0.550             &  & 0.566             & 5476          & 1780             & 95539          & 17514          & 23451             & 575             & 591              \\
      ImmortalTracker~\cite{Wang2021ImmortalTT}           & --              &  & 0.677             & 0.599             &  & \textbf{0.572}    & 5565          & 1669             & 97584          & 18012          & 21661             & 320             & 477              \\
      GNN-PMB~\cite{Liu2022GNNPMBAS}                      & --              &  & 0.678             & 0.560             &  & 0.563             & 5698          & \textbf{1622}    & 97274          & 17071          & 21521             & 770             & 431              \\
      \midrule
      Chiu et al.~\cite{Chiu2020Probabilistic3MM}         & 2D appearance   &  & 0.655             & 0.617             &  & 0.555             & 5494          & \underline{1557} & 95199          & 18061          & 23323             & 1043            & 717              \\
      EagerMOT~\cite{Kim2021EagerMOT3M}                   & 2D geometry     &  & 0.677             & 0.550             &  & 0.568             & 5303          & 1842             & 93484          & 17705          & 24925             & 1156            & 601              \\
      NEBP~\cite{Liang2022NeuralEB}                       & 3D appearance   &  & 0.683             & 0.624             &  & \underline{0.584} & 5428          & 1993             & 97367          & 16773          & 21971             & 227             & 299              \\
      ShaSTA~\cite{Sadjadpour2022ShaSTAMS}                & 3D appearance   &  & \underline{0.696} & 0.540             &  & 0.578             & 5596          & 1813             & 97799          & 16746          & \underline{21293} & 473             & 356              \\
      \midrule
      \name{} (ours)                                      & --              &  & \textbf{0.682}    & \textbf{0.496}    &  & 0.556             & 5466          & 1896             & 95790          & 18322          & 23337             & 438             & 529              \\
      \bottomrule
    \end{tabular}
  \end{center}

  \caption{Results on nuScenes test set using CenterPoint detections. AMOTA and AMOTP are the primary metrics on the benchmark. $^\dagger$denotes offline methods, $^\ddagger$denotes using 10Hz data. We mark best performance in the comparison with baselines in bold text and underline where an even better performance is achieved by methods with additional information besides geometric cues from 3D detections.}
  \label{tab:test_split}
\end{table*}

\begin{table}[t!]
  \centering
  \small
  \setlength{\tabcolsep}{2pt}

  \begin{center}
    \resizebox{\columnwidth}{!}{%
      \begin{tabular}{lccccc}
        \toprule
        Method                                              & AMOTA$\uparrow$                       & AMOTP$\downarrow$ & MOTA$\uparrow$ & IDS$\downarrow$ & FRAG$\downarrow$ \\
        \midrule
        OGR3MOT~\cite{Zaech2021LearnableOG}                 & 0.693 {\scriptsize (+0.037)}          & 0.627             & 0.602          & 262             & 332              \\
        PolarMOT-offline$^\dagger$~\cite{Kim2022PolarMOTHF} & 0.711 {\scriptsize (+0.047)}          & --                & --             & \textbf{213}    & 332              \\
        PolarMOT-online~\cite{Kim2022PolarMOTHF}            & 0.673                                 & --                & --             & 439             & \textbf{285}     \\
        CenterPoint~\cite{Yin2020Centerbased3O}             & 0.665 {\scriptsize (+0.027)}          & 0.567             & 0.562          & 562             & 424              \\
        CBMOT~\cite{Benbarka2021ScoreRF}                    & 0.675 {\scriptsize (+0.026)}          & --                & --             & 494             & --               \\
        SimpleTrack$^\ddagger$~\cite{Pang2021SimpleTrackUA} & 0.696 {\scriptsize (+0.028)}          & 0.547             & 0.602          & --              & 403              \\
        ImmortalTracker~\cite{Wang2021ImmortalTT}           & 0.702 {\scriptsize (+0.025)}          & --                & 0.601          & --              & 385              \\
        GNN-PMB~\cite{Liu2022GNNPMBAS}                      & 0.707 {\scriptsize (+0.029)}          & 0.560             & --             & 650             & 345              \\
        \midrule
        \name{} (ours)                                      & \textbf{0.712} {\scriptsize (+0.030)} & \textbf{0.515}    & \textbf{0.607} & 341             & 436              \\
        \bottomrule
      \end{tabular}}
  \end{center}
  \caption{Results on nuScenes validation set using CenterPoint detections. $^\dagger$ denotes offline methods, $^\ddagger$ denotes using 10Hz data. Changes of AMOTA to the test set are shown in the brackets.}
  \label{tab:val_split}
\end{table}

\begin{table}[t!]
  \centering
  \small
  \setlength{\tabcolsep}{2pt}

  \begin{center}
    \resizebox{\columnwidth}{!}{%
      \begin{tabular}{llccccc}
        \toprule
        Source      & Target      & AMOTA$\uparrow$                      & AMOTP$\downarrow$ & MOTA$\uparrow$ & IDS$\downarrow$ & FRAG$\downarrow$ \\
        \midrule
        CenterPoint & CenterPoint & 0.712                                & 0.515             & 0.607          & 341             & 436              \\
        BEVFusion   & BEVFusion   & 0.749                                & 0.550             & 0.652          & 447             & 443              \\
        MEGVII      & MEGVII      & 0.641                                & 0.639             & 0.535          & 328             & 497              \\
        \midrule
        BEVFusion   & CenterPoint & 0.699 & 0.524             & 0.595          & 421             & 434              \\
        MEGVII      & CenterPoint & 0.697                                & 0.544             & 0.591          & 418             & 445              \\
        CenterPoint & BEVFusion   & 0.747                                & 0.553             & 0.652          & 526             & 475              \\
        MEGVII      & BEVFusion   & 0.744                                & 0.526             & 0.640          & 445             & 479              \\
        CenterPoint & MEGVII      & 0.632                                & 0.643             & 0.529          & 409             & 518              \\
        BEVFusion   & MEGVII      & 0.626                                & 0.663             & 0.521          & 415             & 494              \\
        \bottomrule
      \end{tabular}}
  \end{center}

  \caption{Detector domain generalization experiment. Performance is reported using models that are trained on detections from source detector and tested on detections from target detector.}
  \label{tab:gerneralization}
\end{table}

\section{Experiments}
\label{sec:exp}

\subsection{Experimental Setup}

\paragraph{Dataset}
NuScenes~\cite{Caesar2019nuScenesAM} is a large-scale dataset focusing on perception and prediction for autonomous vehicles that we use for training and testing.
It contains 1000 scenes of 20 second length, which are spilt into 700, 150 and 150 scenes as training, validation and test set, respectively.
The driving data is collected using multiple sensors, including multi-view cameras, a 32-beam LiDAR, RADARs etc.
Despite a higher capture frequency of these sensors, the dataset is annotated at 2Hz.

\vspace{-2mm}
\paragraph{Metrics}
For evaluation, we follow the nuScenes tracking benchmark protocol.
The primary metrics are the AMOTA and AMOTP that are proposed in~\cite{Weng20193DMT}, where AMOTA is used for ranking.
The AMOTA (Average Multi Object Tracking Accuracy) improves the MOTA metric~\cite{Bernardin2008EvaluatingMO} by averaging over the recall-normalized MOTA (MOTAR) over different recall thresholds.
AMOTP (Average Multi Object Tracking Precision) reflects the average of position errors over different recall thresholds.
In addition, nuScenes uses a variety of secondary metrics, \eg MOTA, MOTP, IDS and FRAG from CLEAR MOT~\cite{Bernardin2008EvaluatingMO} and MT/ML from MOT Challenges~\cite{Milan2016MOT16AB,Dendorfer2020MOT20AB}.
These metrics are computed after applying an independent threshold for each class where the highest MOTA is reached.

\vspace{-2mm}
\paragraph{Detector}
As most state-of-the-art methods for MOT report their results on nuScenes using the CenterPoint~\cite{Yin2020Centerbased3O} detector, we use the same detections for a fair comparison.
In addition, we use the detectors from MEGVII~\cite{Zhu2019ClassbalancedGA} and BEVFusion~\cite{Liu2022BEVFusionMM} to validate the generalization of our method.

\vspace{-2mm}
\paragraph{Baselines}
As \name{} only uses 3D detections to accomplish data association, we compare our method with state-of-the-art tracking-by-detection approaches that rely on 3D geometric cues from CenterPoint detections.
Hence, we use the learning-based OGR3MOT~\cite{Zaech2021LearnableOG} and PolarMOT~\cite{Kim2022PolarMOTHF} as primary baselines and additionally
five model-based methods: CenterPoint~\cite{Yin2020Centerbased3O}, CBMOT~\cite{Benbarka2021ScoreRF}, SimpleTrack~\cite{Pang2021SimpleTrackUA}, ImmortalTracker~\cite{Wang2021ImmortalTT}, and GNN-PMB~\cite{Liu2022GNNPMBAS}.

\vspace{-2mm}
\paragraph{Implementation details}
Following SimpleTrack~\cite{Pang2021SimpleTrackUA}, we first use Non-Maximum Suppression (NMS) with a 3D IoU threshold of 0.1 to filter duplicates.
During training, we sample mini-sequences of length $T{=}6$ frames from training scenes as our training samples, which corresponds to 2.5\,s at a frequency of 2Hz.
A track is deleted if it is unmatched for $T_d{=}3$ frames.
All models are trained using AdamW~\cite{Loshchilov2017DecoupledWD} for 12 epochs with a batch size of 8.
We use a learning rate of 0.001 and a weight decay of 0.01.

\subsection{Benchmark Results}

\paragraph{Test set}
\tabref{tab:test_split} shows results on the nuScenes test set using CenterPoint detections.
We first compare \name{} with the baselines that are listed in the upper part of~\tabref{tab:test_split}.
Our approach outperforms other learning-based approaches significantly, yielding 2.6\%P and 1.8\%P AMOTA improvements over OGR3MOT and PolarMOT, respectively.
Furthermore, \name{} surpasses the highest ranking model-based approach GNN-PMB~\cite{Liu2022GNNPMBAS} by 0.4\%P in AMOTA.
In addition, we achieve a notably better AMOTP than the baselines using the same detection boxes, which verifies that our approach can associate more precisely.
Compared to the averaged metrics AMOTA and AMOTP, our best-achieved MOTA at single recall threshold as well as corresponding secondary metrics are relatively lower.
This on the other hand shows a more balanced performance over different recall thresholds of our approach.
Second, we compare to approaches that use additional information besides geometric cues from 3D object detections, \cf the lower half of~\tabref{tab:test_split}.
Our approach still outperforms Chiu \etal~\cite{Chiu2020Probabilistic3MM} and EagerMOT~\cite{Kim2021EagerMOT3M} with 2D data and achieves on par performance to NEBP~\cite{Liang2022NeuralEB}.
Only the concurrent work ShaSTA~\cite{Sadjadpour2022ShaSTAMS} that heavily relies on LiDAR backbone features performs better than our approach.

\paragraph{Validation set}
We further compare to the baselines on the validation set in~\tabref{tab:val_split}, where \name{} again achieves best AMOTA and AMOTP among all approaches.
We observe that the AMOTA of the learning-based baselines (OGR3MOT and PolarMOT) differs more between validation and test split compared to model-based approaches.
This shows some overfitting of learned models, \eg due to hyperparameter tuning using validation performance.
In comparison, the performance difference for \name{} is smaller and close to the one from model-based approaches.
We attribute this observation to our online training strategy that effectively reduces the distribution mismatch between training and inference and hence leads to better generalization, which will be discussed next.

\vspace{-2mm}
\paragraph{Generalization across detectors}
In contrast to typical learning-based approaches, model-based approaches~\cite{Yin2020Centerbased3O,Chiu2020Probabilistic3M,Benbarka2021ScoreRF,Liu2022GNNPMBAS} generalize well across different 3D detectors, as the pre-defined motion models are derived from real world physics and thus independent of the detector.
However, we show that a trained and frozen model of \name{} also generalizes well to different detectors at test time than trained with.
The upper part of~\tabref{tab:gerneralization} shows results, where we use the three different detectors CenterPoint~\cite{Yin2020Centerbased3O}, BEVFusion~\cite{Liu2022BEVFusionMM}, and MEGVII~\cite{Zhu2019ClassbalancedGA} for both training and testing, where \name{} consistently produces high performance for all detectors.
The bottom part shows the results where we train on one source detector and run inference using this trained model on detections from another detector.
For example, when using CenterPoint as target detector, the AMOTA of models trained on BEVFusion and MEGVII detections are only 1.3\%P and 1.5\%P worse than the standard setting.
The same phenomenon can be observed when other detector combinations are used.
Considering that our model can accurately estimate velocities in order to predict the track positions, we account this generalization capability to a learned underlying detector-agnostic motion model using sequential batch optimization.

\vspace{-2mm}
\paragraph{Runtime}
Our approach runs at 54.7 Hz on an Nividia GeForce 2080Ti GPU. It is therefore well suited for real-time applications such as autonomous vehicles.

\subsection{Ablation Studies}
We conduct extensive ablation studies of \name{} to highlight how the proposed components work. All experiments are evaluated on the NuScenes validation set.

\vspace{-2mm}
\paragraph{Training sample length}
We first evaluate the training sample length $T$ which is an important factor for lifting our training to an online setting.
By setting $T=2$, the training degenerates, the autoregressive loop is dropped, and the network becomes an affinity estimator between detections of two frames.
As shown in~\tabref{tab:sample_len}, the performance of $T=2$ is surpassed by larger values with a large margin, \eg 3.8\%P AMOTA for the default setup with $T=6$.
Starting at $T=3$, the training follows the online procedure described in~\secref{sec:online_training} and we observe smaller deltas with increasing $T$.
The results verify the necessity of our proposed training strategy as learning from the whole sequence is critical for the online inference.
The overall performance peaks and converges at $T \in [6,8]$ so we use $T=6$ for all other experiments for training efficiency.

\begin{table}[t]
  \centering
  \small
  \setlength{\tabcolsep}{5pt}

  \begin{center}
    \begin{tabular}{cccccc}
      \toprule
      $T$        & AMOTA$\uparrow$ & AMOTP$\downarrow$ & MOTA$\uparrow$  & IDS$\downarrow$ & FRAG$\downarrow$ \\
      \midrule
      2          & 0.6741          & 0.5738            & 0.5733          & 898             & 520              \\
      3          & 0.7105          & 0.5294            & 0.6095          & 384             & 428              \\
      4          & 0.7096          & \textbf{0.5135}   & 0.6065          & 380             & \textbf{418}     \\
      5          & 0.7111          & 0.5317            & \textbf{0.6115} & 373             & 430              \\
      \textbf{6} & 0.7121          & 0.5149            & 0.6071          & \textbf{341}    & 436              \\
      7          & 0.7120          & \textbf{0.5135}   & 0.6105          & 343             & 432              \\
      8          & \textbf{0.7124} & 0.5203            & 0.6086          & 363             & 434              \\
      \bottomrule
    \end{tabular}
  \end{center}

  \caption{Ablation study on the length of training sample $T$.}
  \label{tab:sample_len}
\end{table}

\begin{table}[t]
  \centering
  \small
  \setlength{\tabcolsep}{2pt}

  \begin{center}
    \resizebox{\columnwidth}{!}{%
      \begin{tabular}{lccccc}
        \toprule
        Variant            & AMOTA$\uparrow$ & AMOTP$\downarrow$ & MOTA$\uparrow$  & IDS$\downarrow$ & FRAG$\downarrow$ \\
        \midrule
        Hungarian Matching & 0.7006          & 0.5245            & 0.5924          & 786             & 492              \\
        GT identity guided & 0.7073          & 0.5200            & 0.6065          & 360             & \textbf{405}     \\
        w/o hidden state   & 0.7013          & 0.5326            & 0.6015          & 371             & 432              \\
        \midrule
        \name{}            & \textbf{0.7121} & \textbf{0.5149}   & \textbf{0.6071} & \textbf{341}    & 436              \\
        \bottomrule
      \end{tabular}}
  \end{center}

  \caption{Ablation study of other training variants.}
  \label{tab:train_opt}
\end{table}

\vspace{-2mm}
\paragraph{Other training-related factors}
We evaluate three different variants that are highly related to our training method:
(1) \textit{Hungarian Matching:} we replace the greedy matching by Hungarian Matching;
(2) \textit{GT identity guided:}
we directly use the annotated association instead of greedy matching to accomplish the track update;
(3) \textit{w/o hidden state:}
we reuse the box embedding as initial track feature instead of using the hidden state from the previous frame $h_{T,i,t-1}^{(L)}$.
As shown in~\tabref{tab:train_opt},
using Hungarian Matching decreases AMOTA by 1.15\%P and introduces a considerable amount of ID switches.
This observation shows that high quality detections should have higher priority in data association, as realized by our greedy strategy that starts with the highest scoring detections.
Guided by GT identity, AMOTA slightly decreases to 0.7073 due to a higher data distribution mismatch between training and inference.
In contrast, our autoregressive forward pass avoids GT information leaking into training data, thus reducing the distribution mismatch.
Without passing track features as hidden state, AMOTA decreases by about 1\%P and the AMOTP is also significantly worse.
This indicates the importance of the hidden state in capturing motion dynamics and the association history.

\begin{table}[t!]
  \centering
  \small
  \setlength{\tabcolsep}{2pt}

  \begin{center}
    \resizebox{\columnwidth}{!}{%
      \begin{tabular}{lccccc}
        \toprule
        Variant         & AMOTA$\uparrow$ & AMOTP$\downarrow$ & MOTA$\uparrow$  & IDS$\downarrow$ & FRAG$\downarrow$ \\
        \midrule
        zero feature    & 0.6469          & 0.6245            & 0.5497          & 3715            & 871              \\
        w/o pred.       & 0.7055          & 0.5315            & 0.5994          & 411             & 445              \\
        w/o time diff.  & 0.7113          & 0.5249            & 0.6108          & 350             & 446              \\
        \midrule
        diff. affinity  & 0.6509          & 0.6185            & 0.5587          & 3547            & 838              \\
        concat affinity & 0.6494          & 0.6170            & 0.5541          & 3308            & 1558             \\
        cosine affinity & 0.6346          & 0.6345            & 0.5417          & 4489            & 922              \\
        \midrule
        \name{}         & \textbf{0.7121} & \textbf{0.5149}   & \textbf{0.6071} & \textbf{341}    & \textbf{436}     \\
        \bottomrule
      \end{tabular}}
  \end{center}

  \caption{Ablation study on different options of edge features. }
  \label{tab:edge_opt}
\end{table}

\begin{table}[t!]
  \centering
  \small
  \setlength{\tabcolsep}{2pt}

  \begin{center}
    \resizebox{\columnwidth}{!}{%
      \begin{tabular}{cccccccc}
        \toprule
                                            & Max dist.         & AMOTA$\uparrow$ & AMOTP$\downarrow$ & MOTA$\uparrow$  & Recall$\uparrow$ &IDS$\downarrow$ & FRAG$\downarrow$ \\
        \midrule
        \multirow{4}{*}{\shortstack[c]{with                                                                                                                     \\ prediction}} & 0.5$\times$ & 0.6952 & 0.5224 & 0.5948 & 0.6952 & 666 & \textbf{313} \\
                                            & \textbf{1.0$\times$} & \textbf{0.7121} & \textbf{0.5149}   & \textbf{0.6071} & 0.7387 & \textbf{341}    & 436              \\
                                            & 1.5$\times$          & 0.7058          & 0.5372            & 0.6026 & 0.7427          & 416             & 517              \\
                                            & 2.0$\times$          & 0.7061          & 0.5384            & 0.6019 & \textbf{0.7509}          & 427             & 539              \\
        \midrule
        \multirow{4}{*}{\shortstack[c]{w/o                                                                                                                      \\ prediction}} & 0.5$\times$ & 0.5822 & 0.6367 & 0.4990  & 0.6423 & 2368 & 942 \\
                                            & 1.0$\times$          & 0.6862          & 0.5505            & 0.5850 & 0.7229          & 1138            & 635              \\
                                            & 1.5$\times$          & 0.7034          & 0.5386            & 0.5975 & 0.7315         & 640             & 500              \\
                                            & 2.0$\times$          & 0.7068          & 0.5362            & 0.5999 & 0.7351           & 447             & 509              \\
        \midrule
        \multicolumn{2}{c}{Fully connected} & 0.7015               & 0.5316          & 0.5937 & 0.7474            & 476             & 493                                \\
        \bottomrule
      \end{tabular}}
  \end{center}

  \caption{Ablation study on dynamic association graph building.}
  \label{tab:graph_building}
\end{table}

\vspace{-2mm}
\paragraph{Edge features}
Next, we show the effectivity of incorporating edge features into our architecture using the Edge-Augmented Graph Transformer~\cite{Hussain2021EdgeaugmentedGT}.
In the upper part of~\tabref{tab:edge_opt}, we analyze the impact of different edge feature embeddings $h_{A}^{(0)}$:
(1) \textit{zero:} edge features are set to zero;
(2) \textit{w/o prediction:} with zeroed value for the center distance after prediction;
(3) \textit{w/o frame difference:} with zeroed value for the frame difference.
Variant \textit{zero} causes a very significant performance drop, which confirms the importance of the edge features.
Variant \textit{w/o prediction} leads to a decrease of AMOTA by 0.66\%P, while \textit{w/o frame difference} has no significant impact on AMOTA.
This observation shows the ability of our model to accurately estimate velocities and to capture the underlying motion model for learning-based data association.
The bottom part shows another variant, where we replace the Edge-Augmented Graph Transformer with a normal Graph Transformer decoder and conduct edge classification using the features of two nodes connected by an edge.
We used the difference, concatenation and the cosine affinity of two node features to estimate the association score.
All three models show a strong performance degradation which again verifies our design choices.

\vspace{-2mm}
\paragraph{Dynamic association graph building}
Besides computational efficiency, the sparsity of the association graph also provides useful structure information for feature interaction and reduces the amount of negative association edges as well as the class imbalance,
as can be seen in~\tabref{tab:graph_building} by comparing our default choice in the second row and a variant with a fully-connected graph in the last row.
In the first part of~\tabref{tab:graph_building}, we vary the class-specific distance threshold for every node in the graph by applying a multiplicative factor.
With 0.5$\times$ distance threshold, many potential connections are missing and this further leads to 1.69\%P AMOTA decrease.
Higher thresholds (1.5$\times$ and 2.0$\times$) increase the recall but on the other hand introduce class imbalance, resulting in about 0.6\%P AMOTA decrease.
In the second part, we use original boxes of tracks without predicting them while constructing the graph.
In this case, AMOTA performance increases with the distance threshold.
Without prediction, tracks have a higher distance to reach the correct association and hence higher thresholds preserve the recall.
However, this setting with prediction performs better and results in a more sparse graph structure.
This again verifies the ability of our approach in estimating velocity accurately to achieve an effective graph building.


\section{Conclusion}
\label{sec:conclusion}
 
In this paper, we presented a novel 3D online multi-object tracking (MOT) framework using Graph Transformer which only relies on geometric cues, termed \name{}.
We formulate the association using a bipartite graph representation and exploit the Edge-Augmented Graph Transformer to reason on the graph structure and conduct data association.
Our network runs recurrently and autoregressively during training and we use a sequential batch optimization to train the network, yielding a fully online training which is closely coupled with the online inference process of MOT.
Our approach achieves state-of-the-art performance on the nuScenes dataset, outperforms all other geometric-based data association approaches, and shows good generalization across different detectors.

\section*{Acknowledgement}

The research leading to these results is funded by the German Federal Ministry for Economic Affairs and Climate Action within the project “KI Delta Learning“ (F\"orderkennzeichen 19A19013A). The authors would like to thank the consortium for the successful cooperation. Juergen Gall has been supported by the Deutsche Forschungsgemeinschaft (DFG, German Research Foundation) GA 1927/5-2 (FOR 2535 Anticipating Human Behavior) and the ERC Consolidator Grant FORHUE (101044724).

\clearpage

\appendix

\section*{\Large Appendix}
\section{Framework Details}
\subsection{Association Graph Representation}
\begin{figure}[t!]
  \begin{center}
    \resizebox{\linewidth}{!}{
      \includegraphics[width=\linewidth]{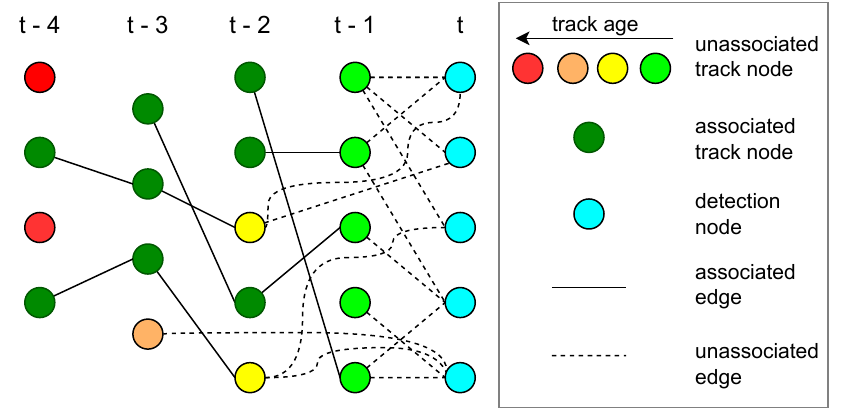}
    }
  \end{center}
  \caption{An illustration of the association graph representation in our framework where each node represents a detection. The graph at time stamp $t$ consists of two sets of nodes that are connected by dashed lines: detection nodes at $t$ (cyan) and track nodes from past frames. To represent the age of a track node, we use a color encoding from light green to red. The associated nodes (dark green) are not processed by the graph.}
  \label{fig:graph}
\end{figure}

We show an illustration of our association graph in~\figref{fig:graph}.
At a time stamp $t$, the association graph is built between detection and track nodes, connected by dashed lines.
The track nodes include inactive nodes that were unassociated at previous frame, \eg yellow and orange nodes at $t-2$ and $t-3$.
If a track node is unassociated for more than $T_d=3$ time stamps, it is permanently deleted from the graph, \eg red nodes at $t-4$.
The association of past frames is shown in solid lines, where associated track nodes (dark green) are removed from the association graph.
As track and detection nodes are two disjoint sets, the association graph is bipartite.
This bipartite representation does not require a complex heuristic algorithm that decodes multi-frame network outputs into hard association.
In contrast, in other approaches that use the spatiotemporal graph with a fixed time window~\cite{Zaech2021LearnableOG,Rangesh2021TrackMPNNAM,Liang2022NeuralEB}, bipartite matching is carried out for every pair of timestamps and hence a conflict resolution step is needed.

\subsection{Track Update Module}
\begin{figure*}[t!]
  \begin{center}
    \resizebox{0.75\linewidth}{!}{
      \includegraphics[width=\linewidth]{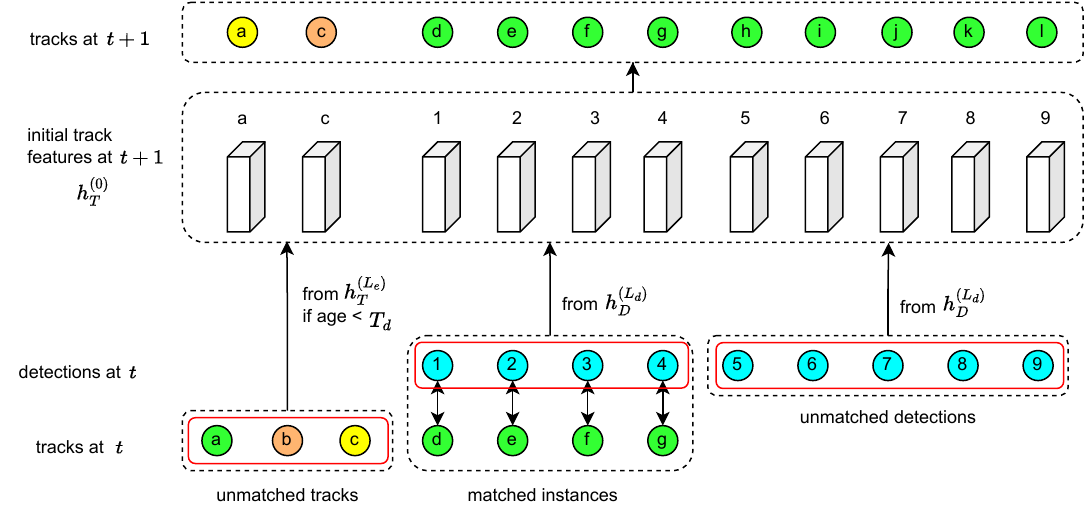}
    }
  \end{center}
  \caption{An illustration of the track update module. Track nodes are indexed by letters in small case and detection nodes in arabic numbers. Color encoding of both track and detection nodes follows~\figref{fig:graph}. The greedy matching generates matched instances, unmatched detections and unmatched tracks. For the three sets, different rules are used to initialize, terminate or update the track. The track node features $h_{T}^{(0)}$ at $t+1$ are initialized from detection $h_{D}^{(L_d)}$ and track features $h_{T}^{(L_e)}$ at time stamp $t$.}
    
  \label{fig:track_update}
\end{figure*}

\figref{fig:track_update} illustrates the details of our track update module.
Given detections $V_D$ and tracks $V_T$ at time stamp $t$ as well as their association score from the network, the greedy matching generates matched track-detection pairs as well as unmatched tracks and unmatched detections.
For matched detection-track pairs, we replace the track features using the feature vectors $h_{D}^{(L_d)}$ of the matched detection instance.
For example, track d is matched with detection 1, therefore the feature vector of detection 1 becomes the feature vector of track d in the next frame. 
Unmatched detections are initialized as new tracks with their features $h_{D}^{(L_d)}$, \eg detection 5 is initialized with track ID h. 
We keep unmatched tracks for $T_d=3$ frames and pass their features $h_{T}^{(L_e)}$ to the next frame, \eg track a and c.
More concisely, this procedure selects the encoder outputs $h_{T}^{(L_e)}$ and decoder outputs $h_{D}^{(L_d)}$ to build track features $h_{T}^{(0)}$ for the next frame, based on the matching results and the rules for spawning and termination.
Besides node features, every track and detection node has additional fields, \eg bounding box parameters, category and velocity, which are used to build the graph in the next frames.
The fields of new track nodes are updated in the same way as the node features.

\subsection{Soft Association}
A drawback of the bipartite graph representation is the limited temporal receptive field.
As shown in~\figref{fig:graph}, the associated nodes (dark green nodes) as well as the associated edges (solid lines) are not included in the graph that will be processed by the Graph Transformer model.

However, this problem is mitigated by the soft association characteristics of transformers.
After being processed by the transformer model, the updated node feature $h_{D,i}^{(L_d)}$ is formed by features from all neighboring track nodes $h_{T,j}^{(L_e)}$ with $j \in \mathcal{N}(i)$ by a weighted average $h_{D,i}^{(L_d)} = \sum_{j \in \mathcal{N}(i)} \alpha_{ij} h_{T,j}^{(L_e)}$, where $\alpha_{ij}$ with $\sum_j \alpha_{ij}=1$ represents the attention learnt by the transformer. 
This updated feature represents an implicit soft association to all combined track nodes, regardless to the hard association decided by greedy matching.
We directly use this feature as initial node features when evolving the graph to the next time stamp $t+1$.
Therefore, the network is aware of the historical information and soft association from last frames.
Combined with our sequential batch optimization during training and back-propagation through time (BPTT)~\cite{Rumelhart1986LearningRB}, the network is trained to be able to correct errors from past frames.

\subsection{Model Details}
Our model employs a feature dimension $d=128$ in all fully connected layers, attention layers, FFNs, \etc.
We stack $L_e=1$ encoder and $L_d=3$ decoder layers.
For all graph self-attention and edge-augmented graph cross-attention layers, we use $C=8$ attention heads and a dropout rate~\cite{Srivastava2014DropoutAS} of 0.1.
Following~\cite{Xiong2020OnLN}, we use LayerNorm~\cite{Ba2016LayerN} in front of all attention and FFN blocks, followed by residual connections~\cite{He2015DeepRL}.

\section{Experiments}

\begin{table}[t!]
  \centering
	\small
	\setlength{\tabcolsep}{2pt}

  \begin{center}
  \resizebox{\columnwidth}{!}{%
	\begin{tabular}{llccccc}
		\toprule
    Detector & Tracker & AMOTA$\uparrow$ & AMOTP$\downarrow$ & MOTA$\uparrow$ & IDS$\downarrow$ & FRAG$\downarrow$\\
    \midrule
    \multirow{4}{*}{MEGVII} & CenterPoint~\cite{Yin2020Centerbased3O} & 0.598 & 0.682 & 0.504 & 462 & 462\\
     & OGR3MOT~\cite{Zaech2021LearnableOG} & 0.631 & 0.762 & \textbf{0.541} & \textbf{263} & \textbf{305} \\
     & GNN-PMB~\cite{Liu2022GNNPMBAS} & 0.619 & 0.716 & -- & 508 & 372 \\
     & \name{} & \textbf{0.641} & \textbf{0.639} & 0.535 & 328 & 497 \\
    \midrule
    \multirow{3}{*}{BEVFusion} & CenterPoint~\cite{Yin2020Centerbased3O} & 0.712 & \textbf{0.542} & 0.616 & 696 & \textbf{346} \\
     & \name{} & \textbf{0.749} & 0.550 & \textbf{0.652} & \textbf{447} & 443 \\
     & CAMO-MOT$^\dagger$~\cite{Wang2022CAMOMOTCA} & 0.760 & 0.561 & -- & 243 & --\\
    \bottomrule
	\end{tabular}}
  \end{center}

	\caption{Comparison with other methods using detections from different detectors on nuScenes validation set. $^\dagger$ denotes method using additional appearance cue for data association.}
  \label{tab:other_detector}
\end{table}

\begin{table*}[htbp!]
  \centering
    \setlength{\tabcolsep}{3pt}
    \def\arraystretch{1.4}

  \begin{center}
    \resizebox{2\columnwidth}{!}{%
    \begin{tabular}{l|lcc|lcccccccccc}
      \hline

      Method & Detector & mAP$\uparrow$ & NDS$\uparrow$ & Tracker & AMOTA$\uparrow$ & AMOTP$\downarrow$ & MOTA$\uparrow$ 
      & IDS$\downarrow$ & FRAG$\downarrow$\\
      \hline
      \multirow{2}{*}{CAMO-MOT} & BEVFusion~\cite{Liu2022BEVFusionMM} \& &70.23 &72.88  & \multirow{2}{*}{CAMO-MOT$^\dagger$~\cite{Wang2022CAMOMOTCA}} & \multirow{2}{*}{0.753} & \multirow{2}{*}{0.472} & \multirow{2}{*}{0.635} & \multirow{2}{*}{324} & \multirow{2}{*}{511}\\
      & FocalsConv~\cite{Chen2022FocalSC} & 63.86 & 69.41 \\
      \hline
      BEVFusion  & BEVFusion-e$^\ddagger$~\cite{Liu2022BEVFusionMM} & 74.99 & 76.09 & CenterPoint Tracker~\cite{Yin2020Centerbased3O} & 0.741 & 0.403 & 0.603 & 506 & 422 \\
      \hline
      MSMDFusion-base & MSMDFusion-base~\cite{jiao2022msmdfusion} & 71.50 &	74.00 & CenterPoint Tracker~\cite{Yin2020Centerbased3O} & 0.740 & 0.549 & 0.624 & 1088 & 743\\
      \hline
      \name{}-BEVFusion & BEVFusion~\cite{Liu2022BEVFusionMM} & 70.23 &72.88 & \name{} (ours) & 0.725 & 0.539 & 0.609 & 593 & 499 \\
      \hline
      TransFusion & TransFusion~\cite{Bai2022TransFusionRL} & 68.90 & 71.68 & CenterPoint Tracker~\cite{Yin2020Centerbased3O} & 0.718 & 0.551 & 0.607 & 944 & 673\\
      \hline
      UVTR-Multimodal & UVTR-Multimodal~\cite{Li2022UnifyingVR} & 67.10 & 71.10 & CenterPoint Tracker~\cite{Yin2020Centerbased3O} & 0.701 & 0.686 & 0.618 & 941 & 798 \\
      \hline
      TransFusion-Lidar & TransFusion-Lidar~\cite{Bai2022TransFusionRL} & 65.52 &	70.23 & CenterPoint Tracker~\cite{Yin2020Centerbased3O} & 0.686 & 0.529 & 0.571 & 893 & 626 \\
      \hline
      \name{}-CenterPoint & CenterPoint~\cite{Yin2020Centerbased3O} & 58.00 & 65.50 & \name{} (ours) & 0.682 & 0.496 & 0.556 & 438 & 529 \\

      \hline
    \end{tabular}}
  \end{center}

    \caption{Results on the nuScenes test set. We compare \name{} using BEVFusion and CenterPoint detections with other tracking-by-detection approaches using different detections in terms of both detection and tracking performance. $^\dagger$ denotes method using additional appearance cue for data association. $^\ddagger$ denotes using model ensemble.}
  \label{tab:test_split2}
\end{table*}

\begin{table}[t!]
    \centering
      \small
      \setlength{\tabcolsep}{2pt}
  
    \begin{center}
      \begin{tabular}{cccccccc}
          \toprule
      $T_d$ & AMOTA$\uparrow$ & AMOTP$\downarrow$ & MOTA$\uparrow$ & FP$\downarrow$ & FN$\downarrow$ & IDS$\downarrow$ & FRAG$\downarrow$\\
      \midrule
      1 & 0.6620 & 0.6495 & 0.5750 & \textbf{10899} & 20395 & 1280 & 1070 \\
      2 & 0.7024 & 0.5577 & \textbf{0.6091} & 11887 & 19413 & 524 & 528 \\
      \textbf{3} & \textbf{0.7121} & 0.5149 & 0.6071 & 13010 & 19281 & 341 & 436 \\
      4 & 0.7121 & 0.4937 & 0.6036 & 14101 & \textbf{19198} & 278 & 388 \\
      5 & 0.7060 & 0.4913 & 0.5956 & 14295 & 19712 & 222 & 363\\
      6 & 0.7046 & \textbf{0.4830} & 0.5974 & 13604 & 20713 & \textbf{195} & \textbf{348} \\
      \bottomrule
      \end{tabular}
    \end{center}
  
      \caption{Ablation study on the maximum track age $T_d$.}
    \label{tab:max_age}
  \end{table}
  
  \begin{table}[t!]
    \centering
      \small
      \setlength{\tabcolsep}{2pt}
  
    \begin{center}
      \begin{tabular}{ccccccccc}
          \toprule
      $d$ & $C$ & $L_e$ & $L_d$ & AMOTA$\uparrow$ & AMOTP$\downarrow$ & MOTA$\uparrow$ & IDS$\downarrow$ & FRAG$\downarrow$\\
      \midrule
      128 & 8 & 1 & 3 & \textbf{0.7121} & \textbf{0.5149} & 0.6071 & 341 & 436 \\
      \midrule
      128 & 8 & 1 & 1 & 0.7065 & 0.5256 & 0.6025 & 371 & 428 \\
      128 & 8 & 1 & 2 & 0.7067 & 0.5225 & 0.6023 & 375 & 421 \\
      128 & 8 & 1 & 4 & 0.7113 & 0.5234 & \textbf{0.6115} & \textbf{327} & 423 \\
      \midrule
      128 & 8 & 0 & 3 & 0.7074 & 0.5288 & 0.6041 & 497 & 434 \\
      128 & 8 & 2 & 3 & 0.7106 & 0.5095 & 0.6070 & 367 & \textbf{420} \\
      128 & 8 & 3 & 3 & 0.7098 & 0.5214 & 0.6046 & 371 & 448 \\
      \midrule
      64 & 8 & 1 & 3 & 0.7084 & 0.5242 & 0.6029 & 360 & 423 \\
      256 & 8 & 1 & 3 & 0.7092 & 0.5201 & 0.6046 & 368 & 422 \\
      \midrule
      128 & 4 & 1 & 3 & 0.7098 & 0.5304 & 0.6112 & 368 & 437 \\
      128 & 16 & 1 & 3 & 0.7087 & 0.5208 & 0.6081 & 372 & 437 \\
      \bottomrule
      \end{tabular}
    \end{center}
  
      \caption{Ablation study on the model hyperparameters.}
    \label{tab:arch}
  \end{table}

\subsection{Comparisons using other Detectors}

To further investigate the generalization on different detectors, we compare the results with other MOT approaches using MEGVII~\cite{Zhu2019ClassbalancedGA} and BEVFusion~\cite{Liu2022BEVFusionMM} detections in~\tabref{tab:other_detector}.
\name{} outperforms OGR3MOT by 1.0\%P AMOTA and achieves the highest AMOTA among all approaches with MEGVII detections.
BEVFusion is published later than our geometry-based baselines and only CAMO-MOT~\cite{Wang2022CAMOMOTCA} evaluated their method using BEVFusion detections. 
We first run the CenterPoint~\cite{Yin2020Centerbased3O} tracking algorithm on BEVFusion detections which achieves a higher AMOTA compared to existing works using CenterPoint detections.
However, our approach again surpasses it by 3.7\%P AMOTA and improves the maximally achievable MOTA by 3.6\%P.
With additional image features for data association, CAMO-MOT achieves an AMOTA improvement of 1.1\%P compared to our approach.

\subsection{Test Results using BEVFusion Detections}
As shown in~\tabref{tab:other_detector}, using BEVFusion detections leads to better MOT performance which we can attribute to its higher detection performance.
To explore the potential of \name{}, we show the tracking results using BEVFusion detections on nuScenes test split and compare with the highest-ranking tracking-by-detection methods in~\tabref{fig:test_summary_bevfusion}.
We also show the mAP (mean Average Precision) and NDS (nuScenes detection score) of the object detectors for all methods because they strongly affect the tracking performance.
Many works~\cite{Liu2022BEVFusionMM,jiao2022msmdfusion,Bai2022TransFusionRL,Li2022UnifyingVR} focus on improving the object detector and use the CenterPoint Tracker~\cite{Yin2020Centerbased3O} for test submission.
As can be observed, the tracking performance of these approaches, especially AMOTA highly depends on the detection performance.
The submission of BEVFusion~\cite{Liu2022BEVFusionMM} uses model ensemble which yields 4.76 \%P mAP and 3.21 \%P NDS improvements compared to a model without ensemble.
However, only a checkpoint without an ensemble of models of BEVFusion is made publicly available for generating detections for tracking. 
In addition, this available checkpoint is trained solely on the training set, whereas the reported 70.23 mAP and 72.88 NDS on the test set are from a model trained on both training and validation set.
This can result in a slightly lower real detection performance on the test set of our used detections than the reported 70.23 mAP and 72.88 NDS.
Despite considerable lower detection performance than BEVFusion-e, \name{} achieves an AMOTA of 0.725 on the nuScenes test split.
Considering the improvements of \name{} against the CenterPoint Tracker in~\tabref{tab:other_detector}, we believe that \name{} can achieve a much higher performance if the unavailable detections of BEVFusion-e are used.
CAMO-MOT~\cite{Wang2022CAMOMOTCA} uses the same BEVFusion checkpoint as ours to generate detections, but augments it with detections from FocalsConv~\cite{Chen2022FocalSC}.
Combined with a tracker based on both geometry and appearance cues, CAMO-MOT achieves the highest 0.753 AMOTA among all methods.

\subsection{Ablation Studies}
We provide more supplementary ablation studies of \name{} to verify our design choices.
All experiments are evaluated on the NuScenes validation set using CenterPoint detections.

\paragraph{Maxixum track age}
The maximum track age $T_d$ is one of the few hyperparameters in the track update module.
In~\tabref{tab:max_age}, we compare the performance of using different values for $T_d$, ranging from 1 to 6.
With $T_d = 1$, unassociated tracks are immediately removed from the graph, which introduces considerable ID switches and fragmentation and finally leads to a significantly lower AMOTA.
The AMOTA peaks at $T_d=\{3, 4\}$  and we use $T_d=3$ as the default setting due to inference efficiency with fewer tracks.
Higher $T_d$ increases the robustness against occlusions or missed detections, but it increases the size of the graph, which makes the training more difficult.
Therefore, when increasing $T_d$, we see a tendency of fewer ID switches and fragmentation, but the AMOTA decreases gradually due to more FPs and FNs.

\paragraph{Model architecture}
We ablate the model-related hyperparameters in~\tabref{tab:arch}, which includes model dimension $d$, number of attention heads $C$, number of encoder layers $L_e$ and decoder layers $L_d$.
Compared to the default setting in the first row, fewer decoder layers causes a drop of AMOTA between 0.5\%P and 0.6\%P.
With $L_e=0$, the tracks from the last frames are directly associated with new detections without updating the track features in advance.
This leads to an AMOTA decrease of 0.47\%P. 
More encoder or decoder layers do not lead to further improvements.
Similarly, varying model dimension and head numbers lead to an AMOTA decrease ranging between 0.2\%P and 0.4\%P.

\paragraph{Loss weights}
We evaluate the impact of the weighting of the loss terms $\mathcal{L}_\text{a}$ and $\mathcal{L}_\text{v}$ by varying the velocity loss weight $\lambda_\text{v}$ in Tab.~\ref{tab:l1_loss}. Using lower or higher weights than $\lambda_\text{v}=1.0$ does not improve AMOTA or AMOTP.

\paragraph{Label assignment}
As shown in Section 3.3, we use Hungarian Matching (HM) with 3D IoU as matching cost to assign object IDs for detections.
However, different ways of label assignment are possible, \eg Greedy Matching (GM), or using other matching costs, \eg center distance.
We show a comparison in Table~\ref{tab:label_generation}. 
Compared to Hungarian Matching, using Greedy Matching decreases the performance slightly.
This observation is different than the track-detection data association that we evaluated in Table 5, where greedy matching performed better.
We argue that the label assignment requires a geometric matching cost like 3D IoU or center distance whereas the data association uses the estimated affinity, thus different matching methods work best.
Results of 3D IoU is better than center distance since it measures the box similarity and not only the location, which results in a preciser lable assignment.

\paragraph{Other factors}
We evaluate four different variants of our approach:
(1) \textit{data augmentation:} the model is trained with random detection dropout and box jitter;
(2) \textit{enc. fully-connected:} we use a standard transformer encoder which utilizes a fully-connected graph;
(3) \textit{w/o velo. estimation:} we remove the velocity estimation of our model, whereas the association graph is built using the velocity from the detector;
(4) \textit{cross entropy:} we replace the focal loss with a binary cross entropy loss.

As can be seen in~\tabref{tab:other_factor}, using data augmentation even results in a performance decrease, which again verifies the generalization ability due to our online training strategy.
A standard transformer encoder leads to an AMOTA decrease of 0.37\%P, which shows the benefit of a sparse graph in the encoder.
\textit{w/o velo. estimation} achieves an AMOTA of 0.7084.
This indicates an accurate enough velocity estimation of the detector for graph building, but our approach estimates velocity more accurately using temporal information of tracked objects.
Cross entropy loss results in an AMOTA decrease of 0.58\%P and this shows the effectiveness of the focal loss for the imbalanced data in our framework.

\begin{table}[t!]
    \centering
    \small
    \setlength{\tabcolsep}{4pt}
    \begin{center}
        \begin{tabular}{lcccccc}
            \hline
            $\lambda_\text{v}$ & 0.2    & 0.6    & \textbf{1.0}    & 2.0 & 5.0\\
            \hline
            AMOTA$\uparrow$     & 0.7078 & 0.7100 & \textbf{0.7121} & 0.7111 & 0.7106 \\
            AMOTP$\downarrow$   & 0.5199 & 0.5217 & \textbf{0.5149} & 0.5317 & 0.5243 \\
            \hline
        \end{tabular}
    \end{center}
    \caption{Ablation study on velocity loss weight $\lambda_\text{L1}$.}
    \label{tab:l1_loss}
\end{table}

\begin{table}[t!]
    \centering
    \small
    \setlength{\tabcolsep}{1pt}

    \begin{center}
        \begin{tabular}{ccccccc}
            \hline
            Matching & Cost         & AMOTA$\uparrow$ & AMOTP$\downarrow$ & MOTA$\uparrow$  & IDS$\downarrow$ & FRAG$\downarrow$ \\
            \hline
            HM       & 3D IoU       & \textbf{0.7121} & \textbf{0.5149}   & \textbf{0.6071} & 341             & 436              \\
            GM       & 3D IoU       & 0.7103          & 0.5244            & 0.6065          & 363             & 429              \\
            HM       & center dist. & 0.7077          & 0.5155            & 0.6045          & \textbf{331}    & 410              \\
            GM       & center dist. & 0.7059          & 0.5190            & 0.6000          & 354             & \textbf{401}     \\
            \hline
        \end{tabular}
    \end{center}
    \caption{Ablation study on matching variants for label generation.}
    \label{tab:label_generation}
\end{table}

\begin{table}[t]
  \centering
  \small
  \setlength{\tabcolsep}{2pt}

  \begin{center}
    \resizebox{\columnwidth}{!}{%
      \begin{tabular}{lccccc}
        \toprule
        Variant         & AMOTA$\uparrow$ & AMOTP$\downarrow$ & MOTA$\uparrow$  & IDS$\downarrow$ & FRAG$\downarrow$ \\
        \midrule
        data augmentation & 0.7092 & 0.5266 & \textbf{0.6076} & \textbf{334} & 440 \\
        enc. fully-connected & 0.7084 & 0.5209 & 0.6016 & 378 & 439 \\
        w/o velo. estimation & 0.7084 & 0.5299 & 0.6121 & 389 & \textbf{430} \\
        cross entropy & 0.7063 & 0.5345 & 0.6059 & 369 & 441 \\
        \midrule
        \name{} & \textbf{0.7121} & \textbf{0.5149} & 0.6071 & 341 & 436 \\
        \bottomrule
      \end{tabular}}
  \end{center}

  \caption{Ablation study on other factors.}
  \label{tab:other_factor}
\end{table}

\subsection{Detailed Metrics of our Test Submissions}
\begin{figure}[p!]
  \begin{center}
    \resizebox{\linewidth}{!}{
      \includegraphics[width=\linewidth]{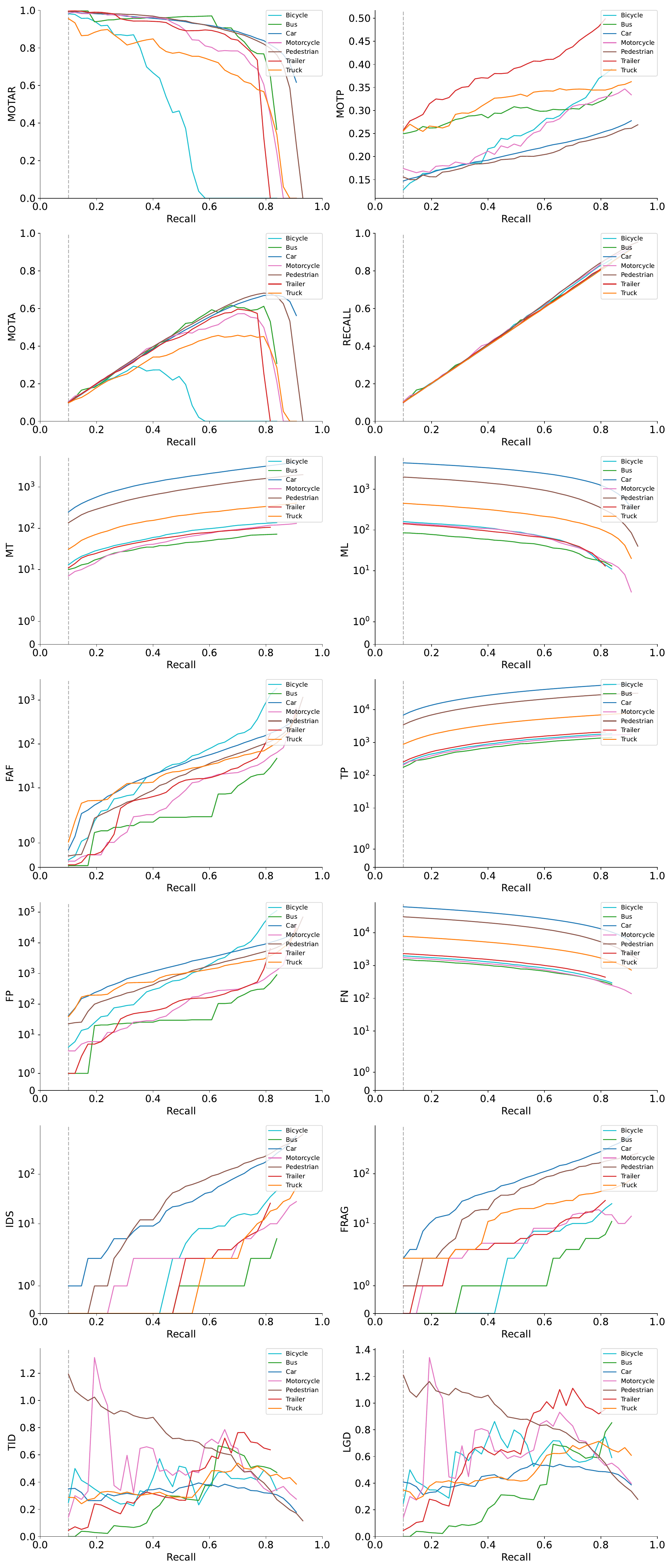}
    }
  \end{center}
  \caption{Test results using CenterPoint detections.}
  \label{fig:test_summary_centerpoint}
\end{figure}

\begin{figure}[p!]
  \begin{center}
    \resizebox{\linewidth}{!}{
      \includegraphics[width=\linewidth]{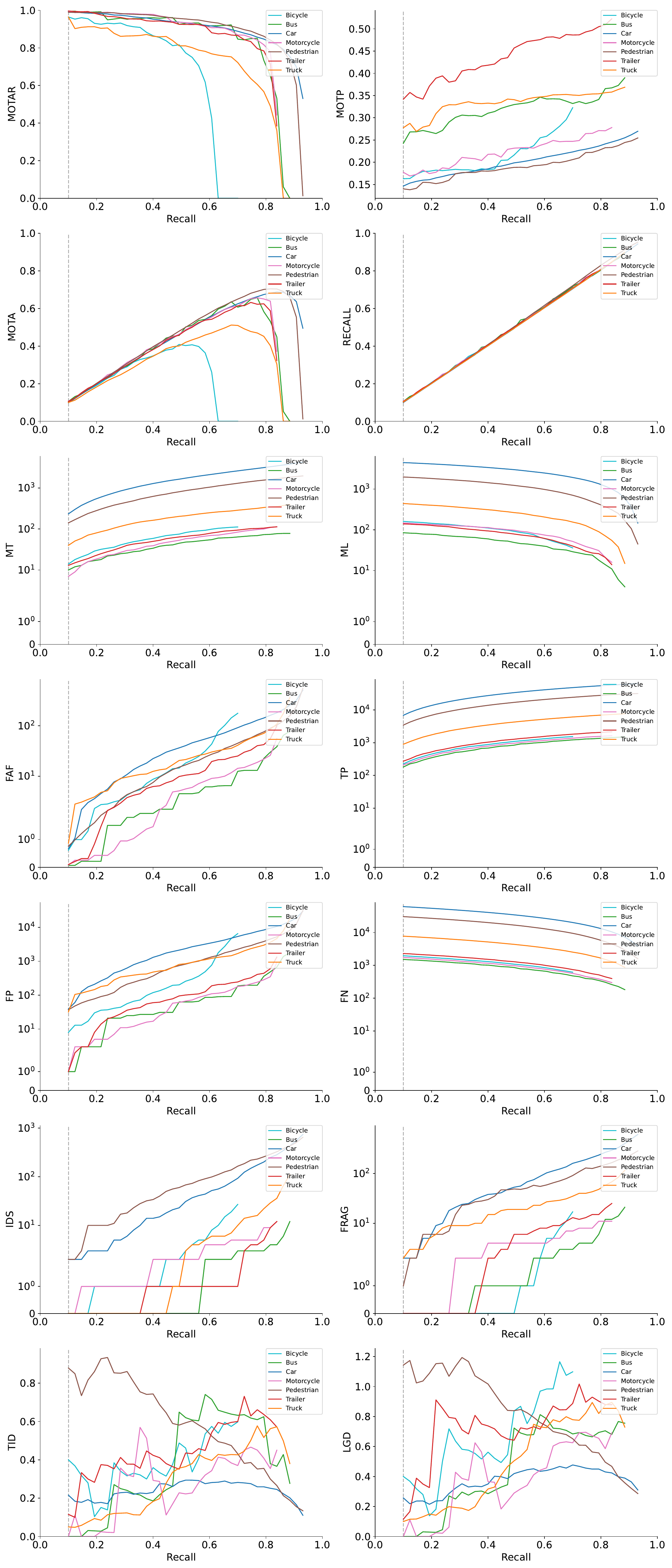}
    }
  \end{center}
  \caption{Test results using BEVFusion detections.}
  \label{fig:test_summary_bevfusion}
\end{figure}

We show the diagrams of different metrics over recalls of our submitted tracking results on the test split.
\figref{fig:test_summary_centerpoint} shows the diagrams of the submission with CenterPoint detections, \figref{fig:test_summary_bevfusion} with BEVFusion detections.
We refer to the nuScenes tracking benchmark website\footnote{benchmark url: \url{https://www.nuscenes.org/tracking?externalData=all&mapData=all&modalities=Any}} for a comparison with other test submissions and a detailed interpretation of all metrics.

\section{Visualization}
\begin{figure}[p!]
  \begin{center}
    frame 12 \\
    \includegraphics[width=\linewidth]{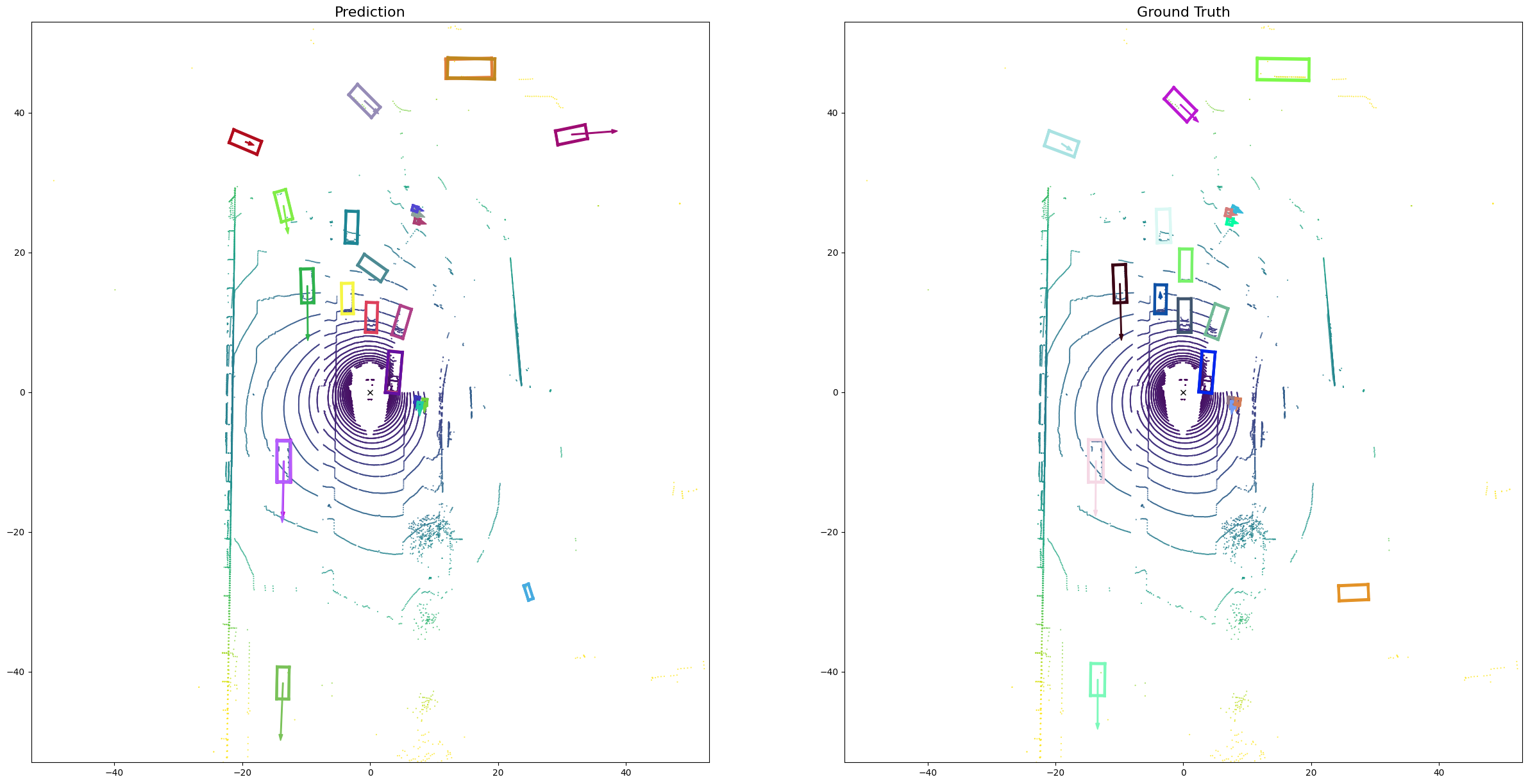}\\
    frame 16 \\
    \includegraphics[width=\linewidth]{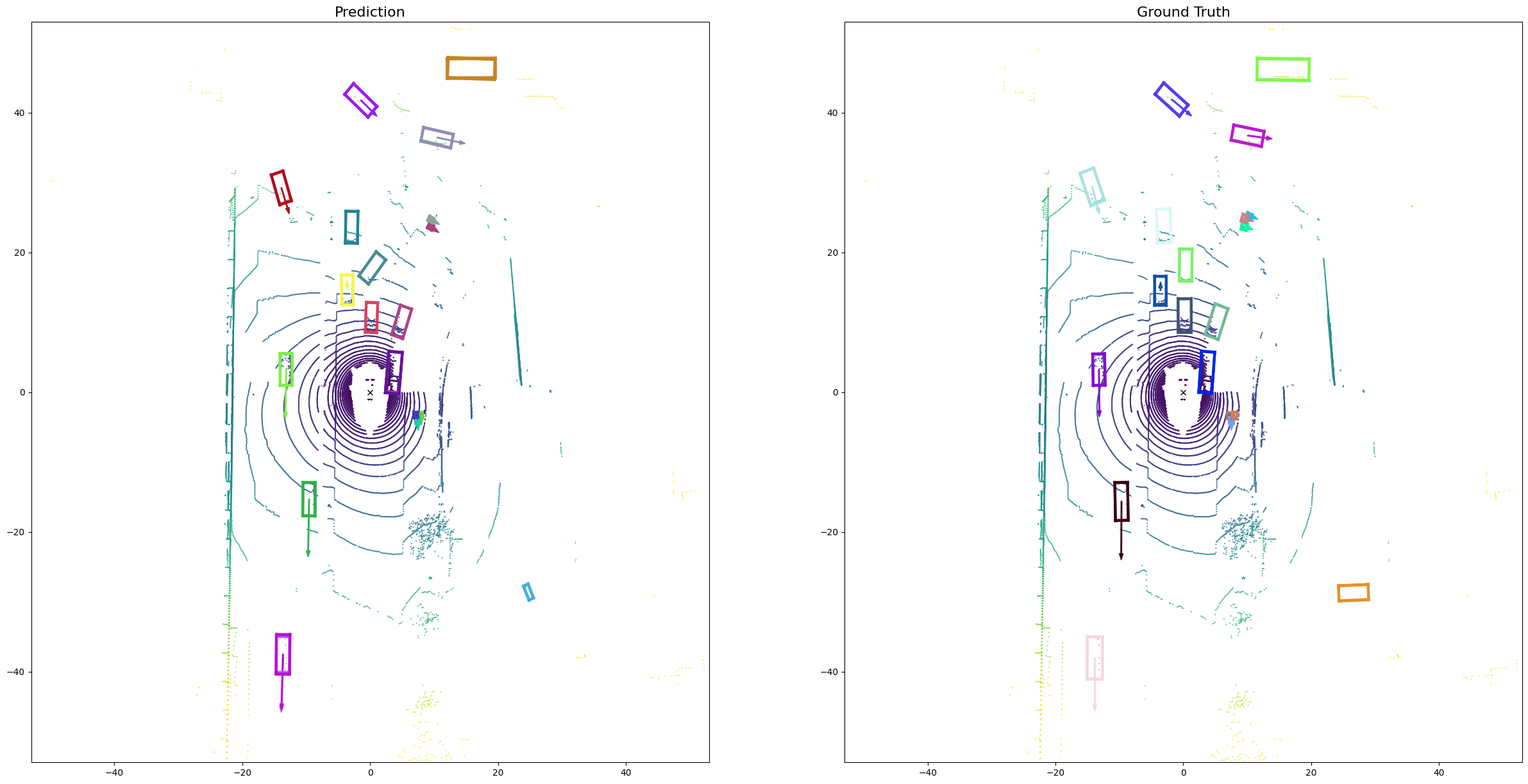} \\
    frame 20 \\
    \includegraphics[width=\linewidth]{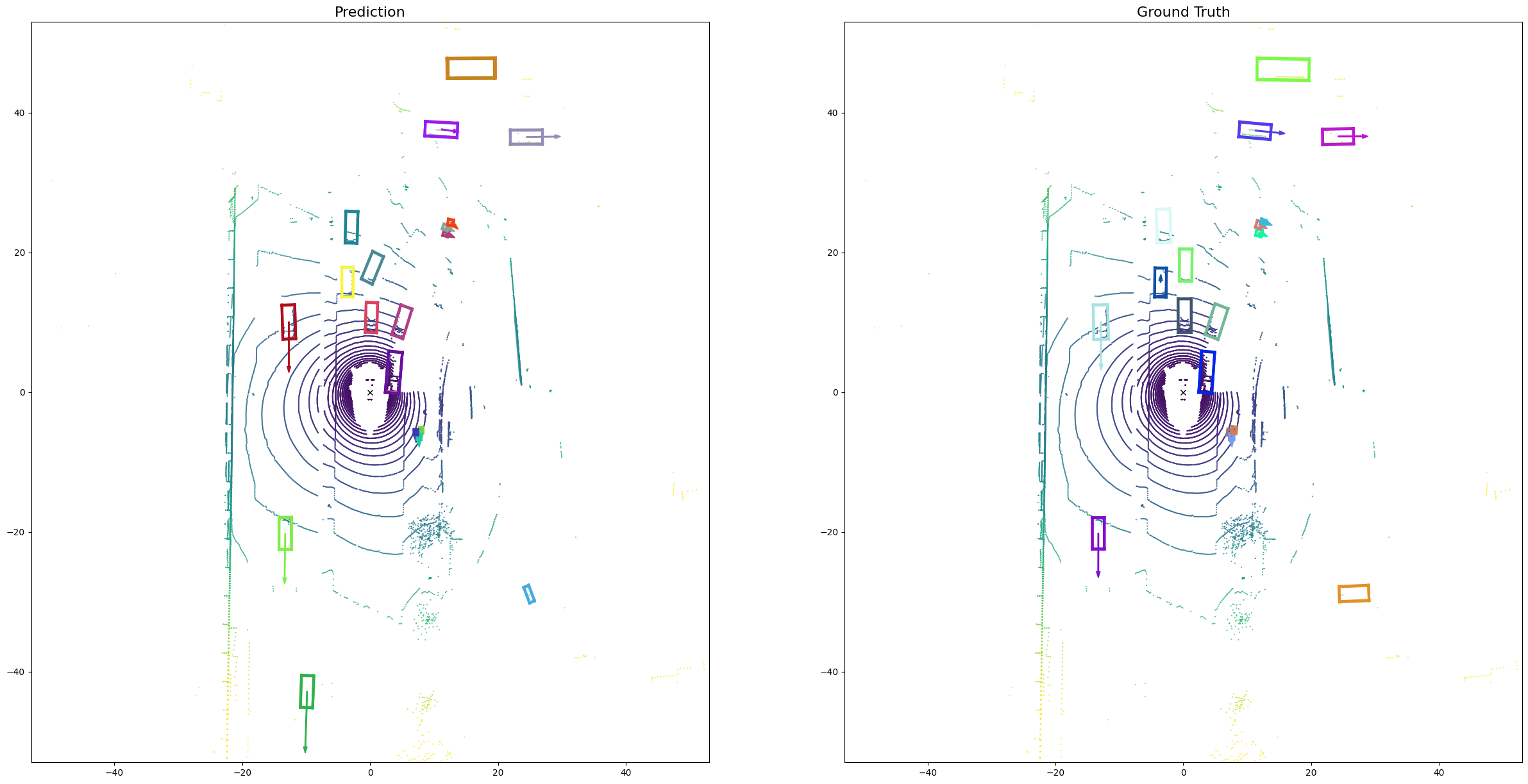} \\
    frame 24 \\
    \includegraphics[width=\linewidth]{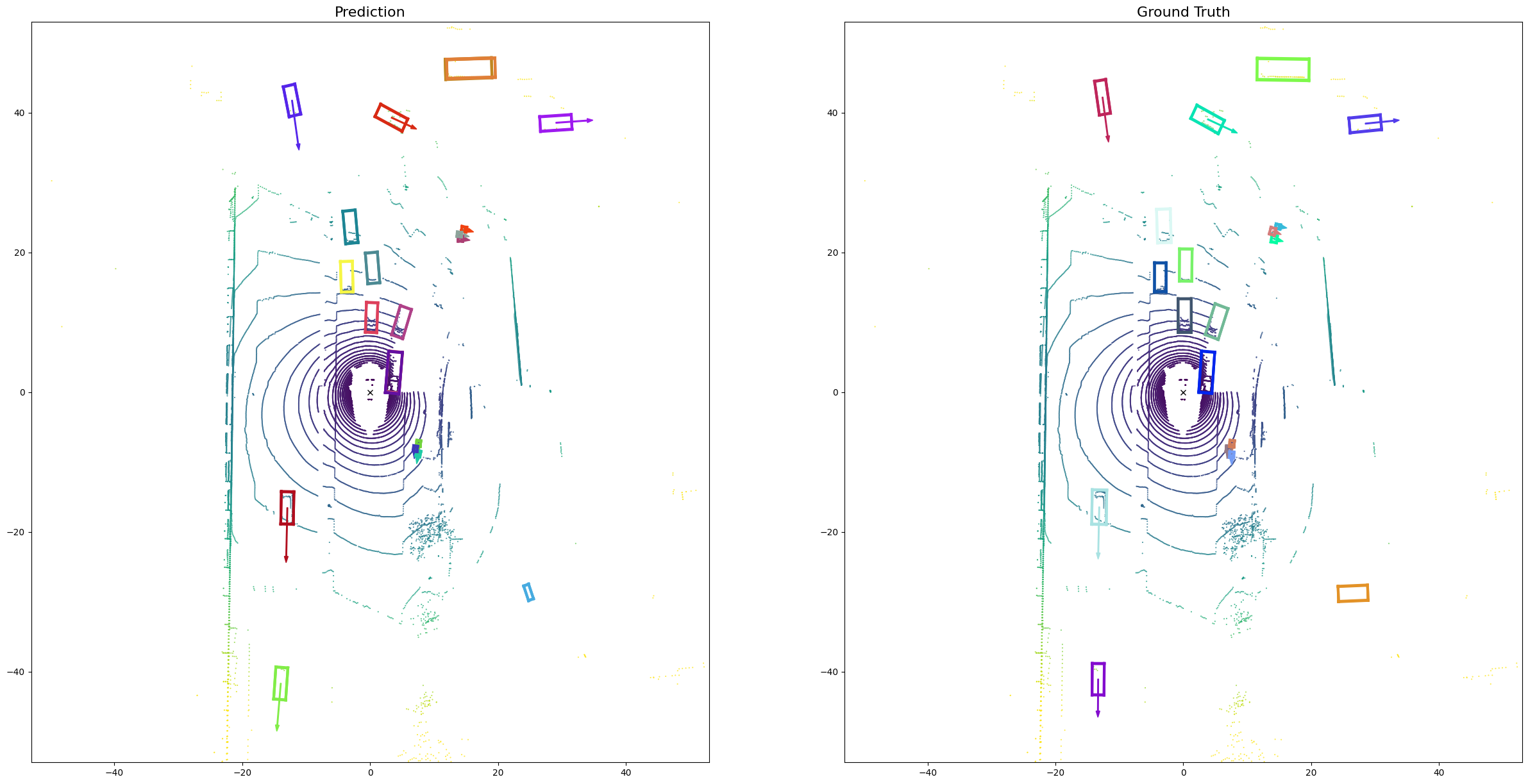} \\
  \end{center}
  \caption{Waiting at intersection.}
  \label{fig:vis1}
\end{figure}

\begin{figure}[p!]
  \begin{center}
    frame 04 \\
    \includegraphics[width=\linewidth]{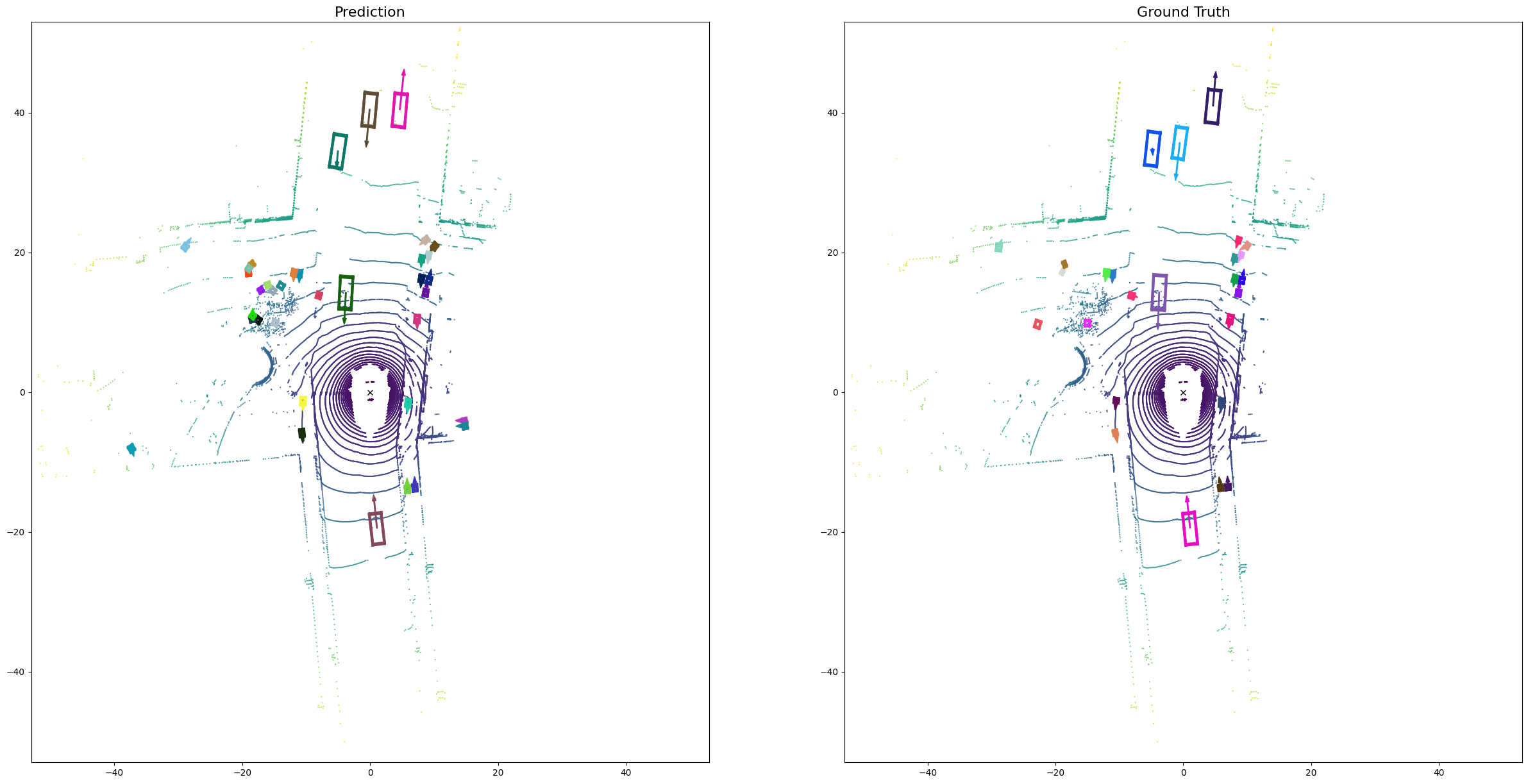}\\
    frame 08 \\
    \includegraphics[width=\linewidth]{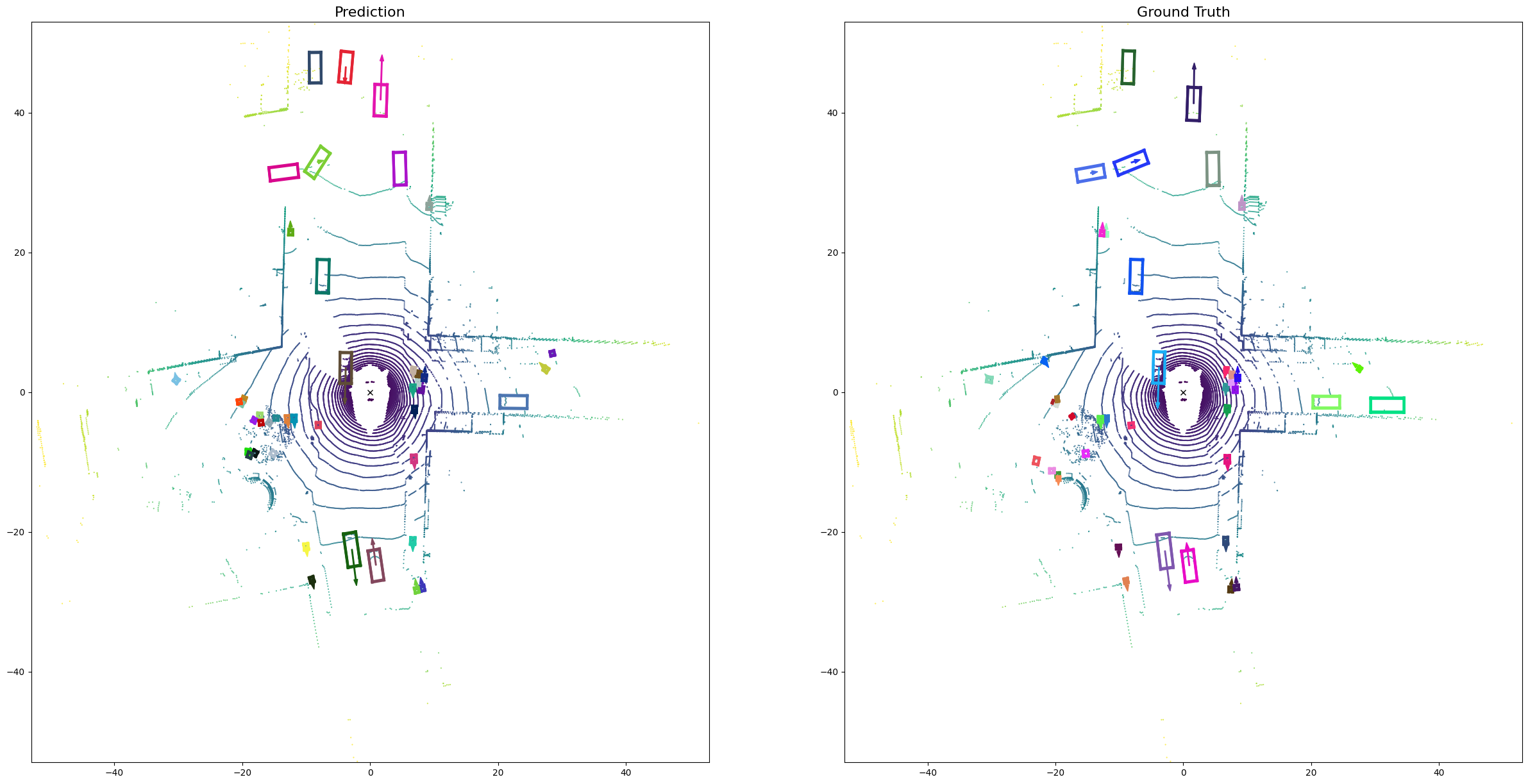} \\
    frame 12 \\
    \includegraphics[width=\linewidth]{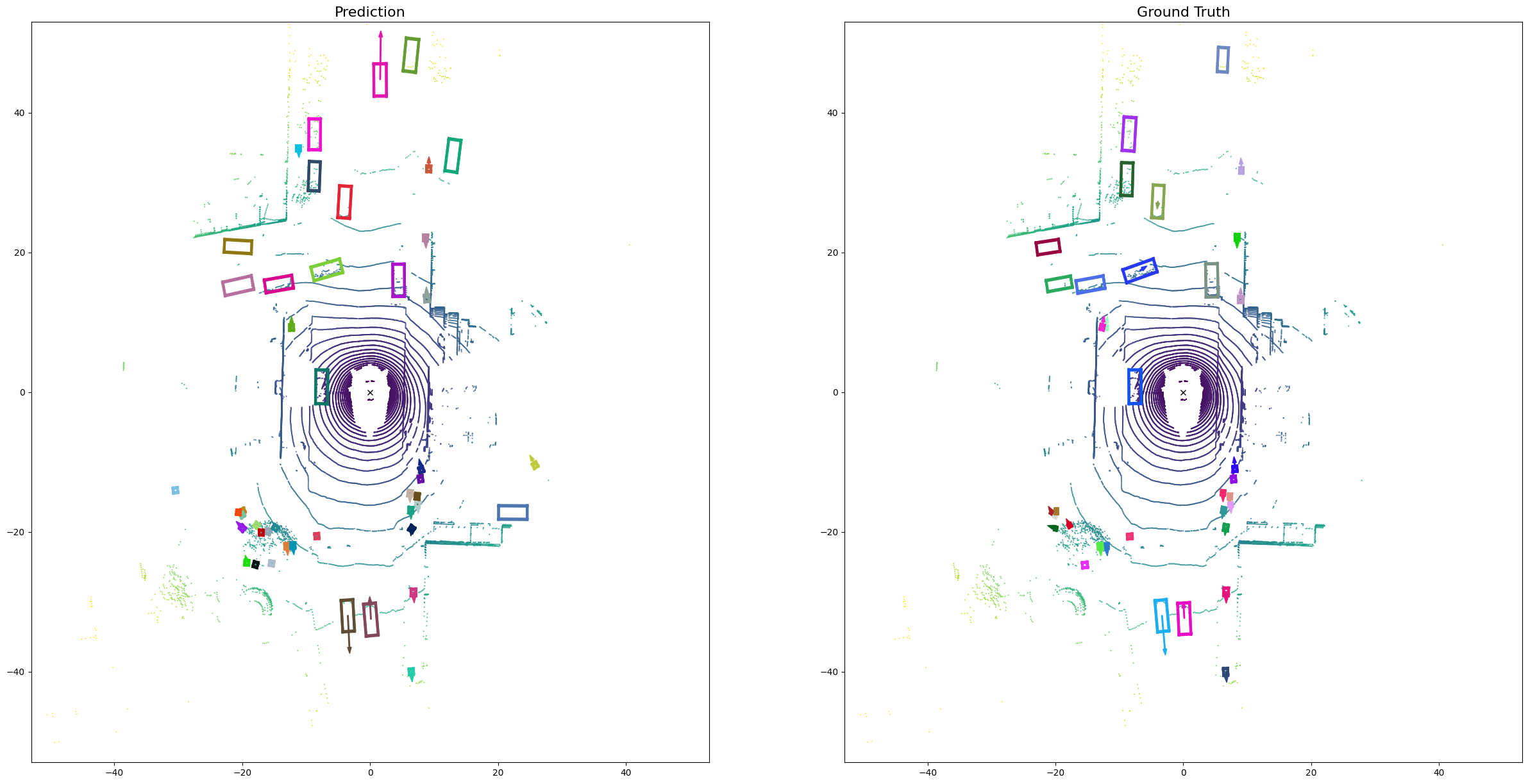} \\
    frame 16 \\
    \includegraphics[width=\linewidth]{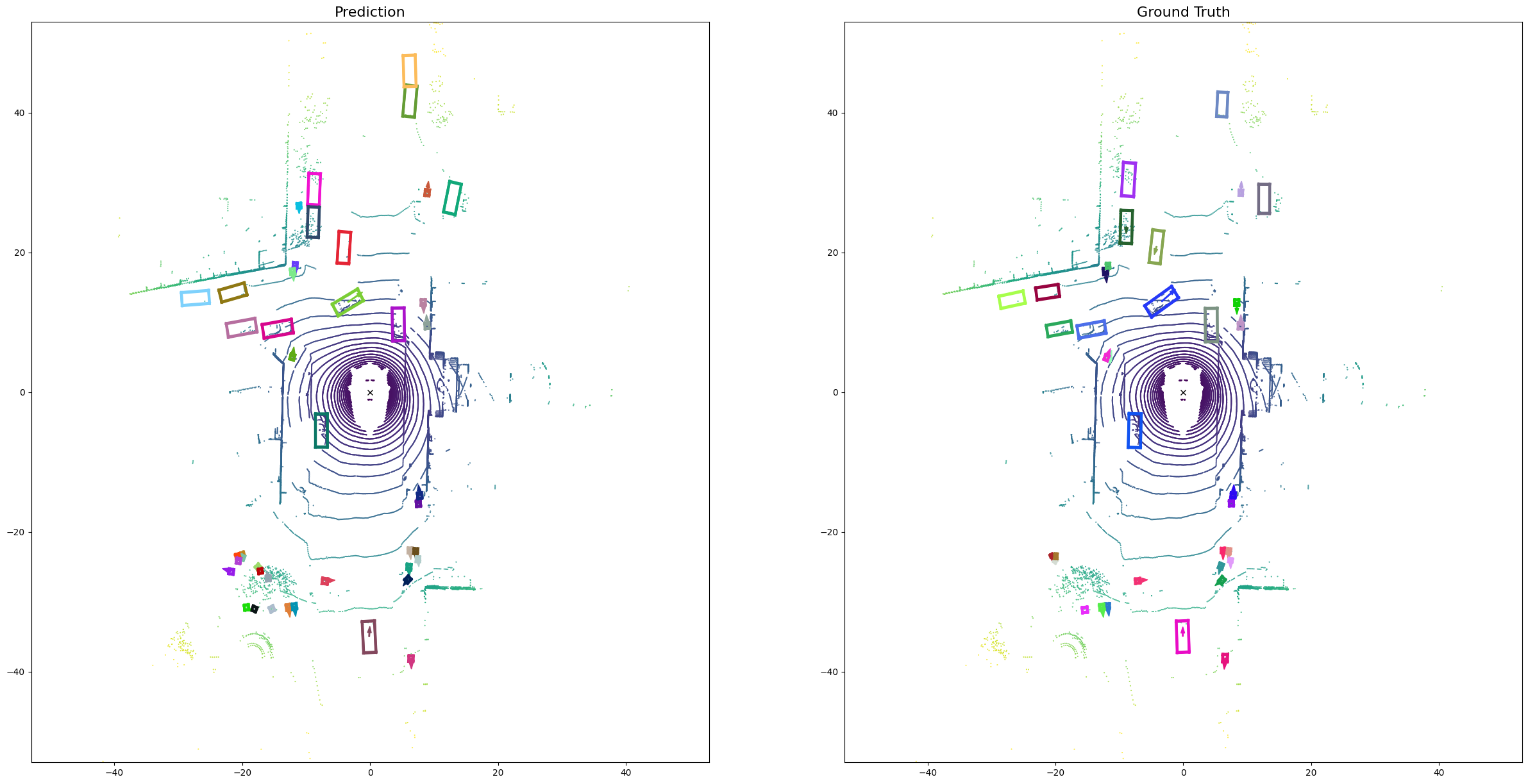} \\
  \end{center}
  \caption{Driving on a street with many traffic participants.}
  \label{fig:vis2}
\end{figure}

We show visualizations of our tracking results (left side) on the nuScenes validation set and compare it with ground truth (right side) in~\figref{fig:vis1} and~\figref{fig:vis2}.
We use a unique color to represent a track ID.
We use arrows to show the velocities of moving objects, where objects with velocity $<$ 0.2 m/s are considered as stationary and no arrows are shown.
\figref{fig:vis1} shows a scenario where the ego vehicle is waiting at a crossing.
We can see a consistent tracking of both moving and static objects as well as an accurate velocity estimation of moving vehicles.
However, an inaccurate orientation of a car in front (frame 12, 16 and 20) or false positives (frame 12) from the object detector can not be corrected by the tracker.
\figref{fig:vis1} shows a scenario where the ego vehicle is moving on a crowded street.
Similarly, we see a consistent tracking, even though cars that are close to each other and extremely small objects such as pedestrians.

\clearpage

{\small
\bibliographystyle{ieee_fullname}
\bibliography{egbib}
}

\end{document}